%% file: Manuscript.tex
\documentclass[smallcondensed]{svjour3}     
\usepackage{pdflscape}

\usepackage{graphicx}
\usepackage{mathtools,nccmath}
\usepackage{adjustbox}
\usepackage{url}
\usepackage{multirow}

\usepackage{subcaption}
\usepackage{arydshln}
\usepackage{algorithm}
\usepackage{algorithmic}
\usepackage{cite}
\usepackage{amsmath,amssymb,amsfonts}
\usepackage{lscape} 
\usepackage{rotating}
\newcommand{\rev}[1]{\color{black}{#1}\color{black}}

\usepackage{graphicx}
\usepackage{textcomp}
\usepackage{xcolor}

\usepackage{balance}

\usepackage{tikz}
\usetikzlibrary{fit,shapes.geometric}

\newcounter{nodemarkers}
\newcommand\circletext[1]{%
    \tikz[overlay,remember picture] 
        \node (marker-\arabic{nodemarkers}-a) at (0,1.5ex) {};%
    #1%
    \tikz[overlay,remember picture]
        \node (marker-\arabic{nodemarkers}-b) at (0,0){};%
    \tikz[overlay,remember picture,inner sep=.5pt]
        \node[draw,ellipse,fit=(marker-\arabic{nodemarkers}-a.center) (marker-\arabic{nodemarkers}-b.center)] {};%
    \stepcounter{nodemarkers}%
}

\newcommand\ourmethod{AdaCC}
\newcommand\nrcompetitors{12}

\usepackage{appendix}

\usepackage{hyperref}
\usepackage{color, colortbl}

\definecolor{Gray3}{gray}{0.6}
\definecolor{Gray2}{gray}{0.7}
\definecolor{Gray1}{gray}{0.8}
\definecolor{Gray0}{gray}{0.9}

\newcolumntype{a}{>{\columncolor{Gray0}}c}
\newcolumntype{b}{>{\columncolor{Gray1}}c}
\newcolumntype{d}{>{\columncolor{Gray2}}c}
\newcolumntype{e}{>{\columncolor{Gray3}}c}

\begin{document}

\title{AdaCC: Cumulative Cost-Sensitive Boosting for Imbalanced Classification}


 \author{Vasileios Iosifidis         \and
         Symeon Papadopoulos \and
         Bodo Rosenhahn \and
         Eirini Ntoutsi
 }


 \institute{V. Iosifidis \at
               Leibniz University of Hannover,
               \email{iosifidis@L3S.de}           
           \and
           S. Papadopoulos \at
               Centre of Research and Technology Hellas,
               \email{papadop@iti.gr}           
           \and
           B. Rosenhahn \at
               Leibniz University of Hannover,
               \email{rosenhahn@tnt.uni-hannover.de}           
           \and
           E. Ntoutsi \at
               Free University of Berlin,
               \email{eirini.ntoutsi@fu-berlin.de}           
 }

\date{Received: date / Accepted: date}

\maketitle

\begin{abstract}
Class imbalance poses a major challenge for machine learning as most supervised learning models might exhibit bias towards the majority class and under-perform in the minority class.
Cost-sensitive learning tackles this problem by treating the classes differently, formulated typically via a \emph{user-defined fixed} misclassification cost matrix provided as input to the learner.  
Such parameter tuning is a challenging task that requires domain knowledge and moreover, wrong adjustments might lead to overall predictive performance deterioration. 
In this work, we propose a novel cost-sensitive boosting approach for imbalanced data that \textit{dynamically} adjusts the misclassification costs over the boosting rounds in response to model's performance instead of using a fixed misclassification cost matrix.
Our method, called \ourmethod{}, is parameter-free as it relies on the \textit{cumulative} behavior of the boosting model in order to adjust the misclassification costs for the next boosting round and comes with theoretical guarantees regarding the training error. Experiments on 27 real-world datasets from different domains with high class imbalance demonstrate the superiority of our method over \nrcompetitors{} state-of-the-art cost-sensitive boosting approaches 
exhibiting consistent improvements in different measures, for instance, 
in the range of [0.3\%-28.56\%] for AUC, [3.4\%-21.4\%] for balanced accuracy, [4.8\%-45\%] for gmean and [7.4\%-85.5\%] for recall.


\end{abstract}

\section{Introduction} 
\label{sec:intro}
\input{introduction}

\section{Related Work}
\label{sec:related}
\input{related}

\section{Preliminaries}
\label{sec:prelim}
\input{preliminaries}

\section{AdaCC: Cumulative Cost-Sensitive Boosting}
\label{sec:method}
\input{method}

\section{Evaluation Setup} 
\label{sec:eval_setup}
\input{evaluation_setup}

\section{Experiments} 
\label{sec:experiments}
\input{experiments}

\section{Conclusions and Future Work} 
\label{sec:conclusions}
\input{conclusions}
\bibliographystyle{spmpsci}      
\bibliography{bibliography}
\newpage

\section*{Appendix}
\input{appendix}

\end{document}

%% file: introduction.tex
When supervised learning models are trained on data generated from skewed class distributions i.e., suffer from the \textit{class imbalance} problem, their performance on  the minority class can degrade significantly, even though they may have outstanding performance in terms of overall error rate or accuracy\footnote{Note: In the binary classification case, the class with significantly more instances is the so-called  \textit{majority} class, while the other is the \textit{minority} class.}.
In extreme cases, the model may ignore the minority class altogether and predict always the majority class. 
 Class imbalance is inherent in many real-world applications e.g., medical diagnosis~\cite{laza2011evaluating,rahman2013addressing}, fraud detection~\cite{phua2004minority,brennan2012comprehensive,sadgaliSB20,di2012improving} or sentiment classification~\cite{li2018imbalanced,iosifidis2019sentiment} and could even lead to discrimination and unfairness~\cite{iosifidis2019fae,iosifidis2018dealing,iosifidis2019adafair,iosifidis2022parity,iosifidis2020mathsf,iosifidis2021online,roy2021multi}.

Over the years, a large body of work has been proposed for tackling the class imbalance problem. 
Following~\cite{sun2007cost}, these works can be categorized into: i) \textit{data-level approaches}, ii)
\textit{model-based approaches}, and iii) \textit{cost-sensitive approaches}. Each category has its own limitations (and strengths). For instance, data-level approaches may discard useful information to restore balance across the different class distributions. 
Model-based approaches are typically designed and implemented for specific models 
and are therefore applicable only in limited settings.
Finally, cost-sensitive methods require as input a misclassification cost matrix thus inducing additional parameters. 

Here, we focus on cost-sensitive classification methods with boosting. 
We have chosen cost-sensitive boosting for three main reasons: i) boosting is able to minimize the training error and at the same time, to avoid overfitting~\cite{schapire1999improved}, ii) boosting is a popular learning method employed in many classification systems \cite{mayr2014evolution}, 
and iii) by re-weighting the data distribution, boosting preserves more information comparing to sampling methods~\cite{sun2007cost}, the prevalent type of data-level methods. 
However, most cost-sensitive boosting methods 
require a fixed misclassification cost matrix provided by the user~\cite{sun2007cost,nikolaou2016cost,fan1999adacost,ting2000comparative}. To define such a matrix, often grid search is performed to find the best costs for the dataset at hand, a tedious and costly process. 
In many cases, as we also show in our experiments, grid search does not lead to optimal selection of misclassification costs. Additionally, having fixed costs during model training may lead to suboptimal learning outcomes. 

\begin{figure*}[htp!]
\centering
  \begin{subfigure}[t]{1\textwidth}
 \includegraphics[width=1\columnwidth, trim=0 0.1cm 0 -1.2cm, clip]{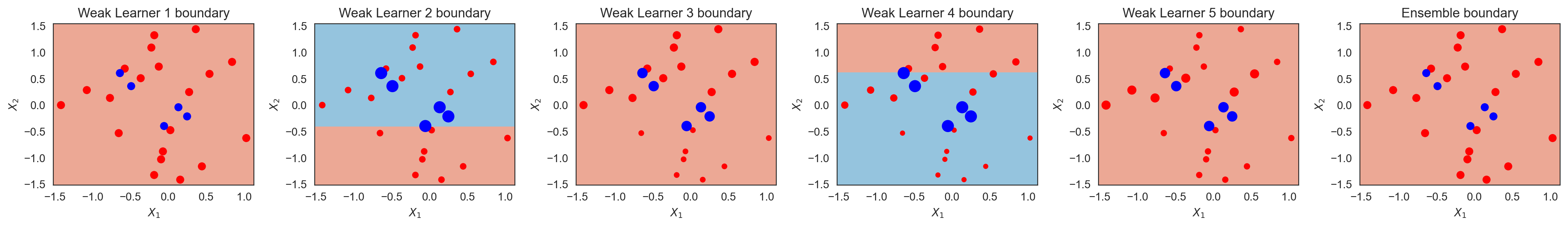}
 \caption{AdaBoost}
 \label{fig:adaboost_toy}
 \end{subfigure}
  \begin{subfigure}[t]{1\textwidth}
  \centering
 \includegraphics[width=1.\columnwidth, trim=0 0.1cm 0 -1.2cm, clip]{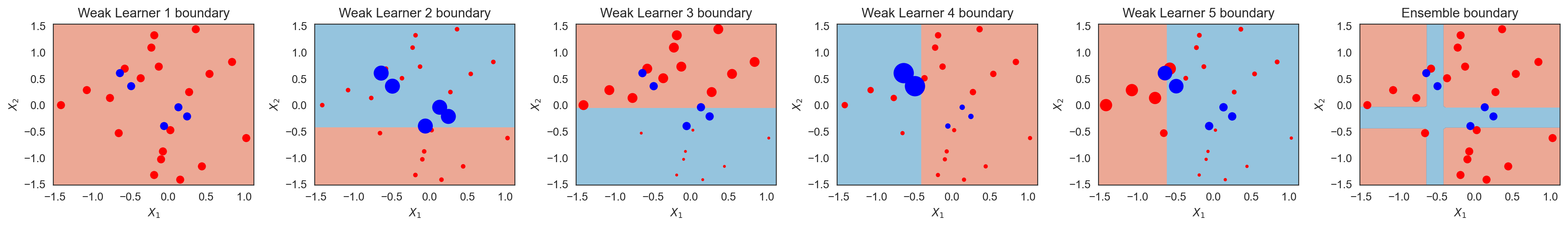}
 \caption{AdaCC1} 
 \label{fig:adacc1_toy}
 \end{subfigure}
   \begin{subfigure}[t]{1\textwidth}
  \centering
 \includegraphics[width=1.\columnwidth, trim=0 0.1cm 0 -1.2cm, clip]{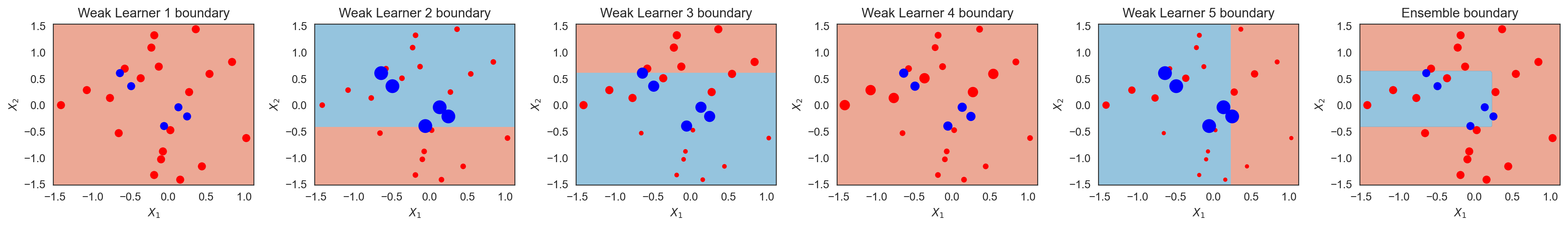}
 \caption{AdaCC2} 
 \label{fig:adacc2_toy}
 \end{subfigure}
 \caption{Decision boundaries of AdaBoost and the two variants of the proposed AdaCC on the same imbalanced toy dataset of 5 blue and 20 red instances. The dot size is proportional to the weight allocated by each learner to the particular instance (with the exception of the last column that depicts the final ensemble), making clear that AdaCC assigns higher weight to minority class instances compared to the ones of the majority class.}
 \label{fig:toy_example}
\end{figure*}


In this work, we propose a new parameter-free cost-sensitive boosting approach for classification problems with high class imbalance. The proposed method, named \emph{AdaCC}, standing for \emph{Cumulative Cost-Sensitive Boosting}, alleviates the need for setting a fixed misclassification cost-matrix as input parameter, by leveraging the cumulative costs of the model up to the current boosting round. As we show in Section \ref{sec:method}, the proposed method has proven upper bounds for the training error. We propose two variants of the method, \textit{AdaCC1} and \textit{AdaCC2} that differ in terms of the employed data re-weighting scheme.

\rev{We carry out a comprehensive experimental study on 27 real world datasets and compare our method with \nrcompetitors{} state-of-the-art cost-sensitive boosting methods as well as 5 non cost-sensitive class-imbalance methods. Our results demonstrate the superior performance of AdaCC over the state of the art in terms of AUC, balanced accuracy, geometric mean, and recall. } 
Notably, the performance improvements are more pronounced on the minority class. This makes our method suitable for tasks where high false negative rates are critical, e.g., medical diagnosis, fraud detection, fairness-aware machine learning, etc. 



 \rev{Figure~\ref{fig:toy_example} illustrates a binary imbalanced toy dataset where the blue points (\#5) belong to the minority class and the red points (\#20) to the majority class. We compare the learning behaviour of the proposed method with the one by AdaBoost by observing the decision boundaries of the weak learners on the toy dataset as well as the ensemble boundary at the end of the training process (rightmost figure). 
Due to the low dimensionality of the dataset and for illustration purposes, we use a small number of $T=5$ weak learners. Figure~\ref{fig:toy_example} demonstrates how weighted data distribution is affected by each weak learner as well as the decision boundary of the final ensembles. The first 5 columns correspond to the 5 weak learners while the size of the dots corresponds to the weight that each instance receives per round. The last column illustrates the decision boundary of the ensemble model. } We note that the final AdaCC model manages to fit  better the data distribution compared to AdaBoost, allocating a ``proper'' part of the feature space to the minority class.

The rest of the paper is organized as follows: Related work is summarized in Section~\ref{sec:related}. Basic concepts are described in Section~\ref{sec:prelim}. Our approach is introduced in Section~\ref{sec:method}. Evaluation setup and experimental results are presented in Sections~\ref{sec:eval_setup} and~\ref{sec:experiments}, respectively. Finally, Section~\ref{sec:conclusions} concludes our work and identifies interesting directions for future research.

%% file: related.tex
Methods for dealing with class imbalance can be organised in three broad categories~\cite{sun2007cost}: i) data-level, ii) model-based and iii) cost-sensitive methods.

\textbf{Data-level methods}
 operate at the dataset level i.e., they modify the data distribution before model training, 
making these methods universally applicable. In~\cite{japkowicz2002class}, the authors investigate the problem of class imbalance and the impact of re-sampling methods under the inter-dependencies of class distribution skewness, data complexity, data volume and employed models. In~\cite{ling1998data} the authors propose a combination of under- and over-sampling to equalize class distributions and measure  model's performance using lift analysis. The impact of over-sampling and under-sampling under the cost curves performance metrics has been explored in~\cite{drummond2003c4}. The authors conclude that under-sampling is significantly more effective than over-sampling for C4.5 classifiers. 
In~\cite{chawla2002smote}, the authors propose \textit{SMOTE}, a method that augments the minority class by interpolating new instances in local neighborhoods.
In~\cite{iosifidis2019sentiment}, the authors propose text augmentation techniques, such as distortion and semantic similarity, to increase the representation of the minority class.

Although re-sampling approaches are simple and easy to use, they come with disadvantages. For example, over-sampling may fail to ``boost'' existing rare cases, and adds no additional information to the dataset~\cite{drummond2003c4,sun2007cost}. Under-sampling on the other hand, can deteriorate the performance by removing important information from the majority class~\cite{chawla2002smote}. Finally, augmentation methods can amplify and propagate noise~\cite{iosifidis2019sentiment}, leading to overall performance detoriation.

\textbf{Model-based methods} tackle class imbalance during training either by employing a mechanism which aims to identify rare patterns or by optimizing for a balanced-performance aware metric. \textit{SMOTEBoost}~\cite{chawla2003smoteboost} combines SMOTE~\cite{chawla2002smote} and AdaBoost~\cite{schapire1999brief} to deal with class imbalance by augmenting the minority class in each boosting round. A similar line of work is \textit{RUSBoost}~\cite{seiffert2009rusboost}, which combines AdaBoost and random under-sampling of the majority class on each boosting round. 
\textit{DataBoost-IM}~\cite{guo2004learning} locates the hard-to-learn instances from both positive and negative classes during the training phase of AdaBoost and based on these instances, it generates synthetic data for augmentation at the end of each boosting round. 
 Class imbalance-sensitive pruning of decision trees has been presented in~\cite{zadrozny2001learning}. The work in~\cite{wu2003class} uses a kernel alignment to optimize the decision boundary of an SVM.  In \cite{krasanakis2017}, a class posterior re-balancing framework has been proposed to reduce imbalance while retaining  classification certainty.
Over the recent years, hybrid methods have also been proposed. In~\cite{wu2017multiset}, they employ multi-set feature learning to learn discriminant features from the constructed multi-set and combine the sets with a generative adversarial network technique such that each subset has similar distribution with the original dataset. In~\cite{yin2020novel}, authors propose a combination of different techniques such as under/over-sampling, data transformations, misclassification costs and ensemble learning to deal with class imbalance. 

The main disadvantage of model-based methods is that the inductive bias of the selected model can raise issues given an imbalanced dataset e.g., decision tree's data fragmentation problem~\cite{he2013imbalanced}.  Additionally, they typically rely on assumptions regarding the underlying data properties or are tailored to specific classification algorithms, which makes hard their application to new domains and datasets. 

\textbf{Cost-sensitive methods} 
do not optimize for overall accuracy. Instead, they try to minimize the overall misclassification costs. This class of algorithms is divided into three sub-categories~\cite{sun2007cost}: i) weighting the data space, ii) making a specific classifier cost-sensitive, and iii) using the Bayes risk theory to assign each instance to the class with the lowest risk. The first sub-category aims to alter data distribution by employing a misclassification cost matrix such that errors in  minority class instances induce a higher loss.
The very first method in this line of work is \textit{AdaCost}~\cite{fan1999adacost}. Over the years many variations of AdaCost have been introduced such as: \textit{CSB1}~\cite{ting2000comparative}, \textit{CSB2}~\cite{ting2000comparative}, \textit{RareBoost}~\cite{joshi2001evaluating}, \textit{AdaC1}~\cite{sun2007cost}, \textit{AdaC2}~\cite{sun2007cost}, \textit{AdaC3}~\cite{sun2007cost} and \textit{CGAda}~\cite{landesa2012shedding,landesa2015revisiting_a,landesa2015revisiting_b}, which differ in the following main aspects: training data weight assignments, weight update rules, and decision rules.
Except RareBoost, all the aforementioned methods in this category require user parameters for the misclassification costs.  An overview of these methods can be seen in Table~\ref{tbl:baselines}.
The second sub-category of cost-sensitive methods aims to make a specific classifier cost-sensitive. In~\cite{nikolaou2015calibrating}, authors propose \textit{AdaMEC}, a boosting classifier that uses the misclassification costs only to set thresholds to the decision boundary of AdaBoost, in contrast to the previous methods which use the misclassification costs to change the data distribution in each boosting round. CGAda and AdaMEC have also been extended in~\cite{nikolaou2016cost}, namely \textit{CGAda-Cal.} and \textit{AdaMEC-Cal.}, by calibrating the models' scores using the Platt scaling technique~\cite{platt1999probabilistic}. In~\cite{qin2013cost}, a cost-sensitive $k$-NN classifier is introduced to tackle class imbalance by using a modified distance function which takes into consideration the misclassification cost matrix. In~\cite{ling2004decision}, a misclassification cost matrix is used to define a cost-sensitive splitting criterion in decision trees, while in~\cite{bradford1998pruning} the authors take into account the misclassification costs to determine the pruning criterion of a decision tree. The third sub-category uses the Bayes risk theory to assign each instance to a class with the lowest risk. Few works have been proposed in this direction e.g., in~\cite{domingosgeneral} the authors swap the class labels of the leaves to minimize the misclassification cost.

For evaluation purposes, we select all the aforementioned cost-sensitive boosting methods since they are related to our contribution. In contrast to our proposed approach, however, the aforementioned cost-sensitive boosting methods assume that the misclassification costs for each class are known in advance (except RareBoost). For many applications/datasets these costs might not be available, and a costly grid search has to be performed to estimate them; however, in many cases, even grid search does not lead to optimal misclassification costs. Instead, the two variants of our approach are parameter-free and leverage the cumulative behavior of AdaBoost to dynamically adjust the misclassification costs per boosting round. Hence, our methods are applicable to any imbalanced dataset without any prior domain knowledge.

%% file: preliminaries.tex
\rev{For the sake of clarity, in Table~\ref{tbl:notation} we briefly describe the employed notations. } We assume a set of instances $D=\{(x_1,y_1),\cdots,(x_n,y_n)\}$ consisting of $n$ independent and identically distributed samples drawn from the joint distribution $P(A,y)$, where
$A$ denotes the feature space and $y$ is the class attribute. For simplicity, we assume the class is binary with $y \in \{+1, -1\}$.
We denote by $D_+$ ($D_-$) the set of instances belonging to the positive (negative, respectively) class. We also assume that 
the positive class is the minority, i.e., $|D_+| << |D_-|$.
It holds that  $|D_+| + |D_-| = n$.

\begin{table}[t!]\centering
\caption{Notations}
\label{tbl:notation}
\begin{tabular}{ll}
\hline
$D$ & set of instances \\  
$x$ & sample \\
$A$ & feature space \\ 
$y$ & class attribute \\
$\vec{C}$ & misclassification cost vector \\
$T$ & number of weak learners \\ 
$h_t$  & weak learner, at round $t$ \\ 
$\alpha_t$ & weight of the weak learner, at round $t$ \\ 
$Z_t$ & normalization factor, at round $t$   \\ 
$H(\cdot)$ & ensemble  \\ 
$sign(\cdot)$ & sign function  \\ 
$exp(\cdot)$ & exponential function  \\ 
$\mathbb{I}\left\{\cdot \right\}$ & Indicator function\\
\hline
\end{tabular}
\end{table}

Standard classification models treat instances of different classes equally and the performance of the induced classifier (see confusion matrix in Table~\ref{tbl:conf_matrix}) is measured in terms of the overall error rate (ER) as: $ER = (FP + FN)/(TP + TN + FP + FN)$. However, when the class distribution is skewed, the overall error rate is not a good indicator of model's performance in all classes, but rather of the performance on the majority class.
In such a case, more apprioprate  performance metrics should be employed (see an overview in Table~\ref{tbl:measures}).

Cost-sensitive models tackle the class imbalance problem by emphasizing more on the minority class  through appropriate costs~\cite{nikolaou2016cost,he2013imbalanced,ting2000comparative}. Each sample $x \in D$ is mapped to a typically \emph{fixed} misclassification cost vector $\vec{C}=<C_+,C_->$, where each sample in $D_+$ is associated with a fixed cost value $C_+$ from the misclassification cost vector $\vec{C}$ and each sample in $D_-$ with a fixed cost value $C_-$ from $\vec{C}$, where $C_+ > C_-$ and $C_+,C_-\in[0,\infty)$.
The costs denote the misclassification costs for each class and are employed by the cost-sensitive learner during the  training phase to ``force'' the learner to also learn minority instances.
The costs, however, need to be manually set by the user, thus requiring prior domain knowledge, or to be selected via grid search~\cite{sun2007cost,he2013imbalanced}.

\begin{table}[t!]\centering
\caption{Confusion Matrix.}
\label{tbl:conf_matrix}
\begin{adjustbox}{width=0.8\columnwidth,center}
\begin{tabular}{lcc}
\hline
         & \textbf{Predicted Positives}  & \textbf{Predicted Negatives}  \\ \hline
\textbf{Positives} $(|D_+|)$ & True Positives (TP)  & False Negatives (FN) \\
\textbf{Negatives} $(|D_-|)$ & False Positives (FP) & True Negatives (TN)  \\ \hline
\end{tabular}
\end{adjustbox}
\end{table}

\textbf{Boosting and AdaBoost:} 
Boosting is an ensemble learning technique which trains a sequence of $T$ weak learners, in order to create a strong learner. The sequential generation  promotes the dependency between the weak learners and each learner learns from the mistakes of the previous learner.

AdaBoost~\cite{schapire1999brief}, one of the most popular boosting algorithms (see Algorithm~\ref{alg:adaboost}),
adjusts
in each iteration $t:1-T$ (the so-called \textit{boosting round} $t$) the data distribution $D^t$ based on the mistakes of the current learner $h_t$ in order to focus in the next round $t+1$ on the misclassified instances. In particular, the weights of the instances for the next round are updated as follows: 
\begin{equation}
\label{eq:adaboost_d}
D^{t+1}(i) = \frac{D^t(i)\exp{(-\alpha_ty_ih_t(x_i))}}{Z_t}
\end{equation} 
The parameter $\alpha_t$ denotes the weight of the weak learner $h_t$ in the final classification decision and is based on the error rate of the weak learner $h_t$: 
\begin{equation}
\label{eq:adaboost_a}
\alpha_t = \frac{1}{2}\log\left(\frac{\sum\limits_{i,y_i=h_t(x_i)} D^t(i)}{\sum\limits_{i,y_i\ne h_t(x_i)} D^t(i)}\right)
\end{equation} 
The parameter $Z_t$ is a normalization factor which is used at the end of each boosting round to make $D^{t+1}$ a probability distribution:
\begin{equation}
\label{eq:Z_adaboost}
Z_t = \sum\limits_{i=1}^n D^t(i)exp\left(-\alpha_ty_ih_t(x_i)\right)
\end{equation}
The final model is a weighted combination of the weak learners:
\begin{equation}
\label{eq:H_adaboost}
H(x)=\text{sign}\left(\sum\limits_{t=1}^T \alpha_th_t(x)\right)
\end{equation}

\begin{algorithm}[tp!]
  \caption{AdaBoost Algorithm}
\label{alg:adaboost}
 \begin{algorithmic}[1]
 \renewcommand{\algorithmicrequire}{\textbf{Input: $D = (x_i,y_i)_1^n$, where $y \in \{+1,-1\}$, $T$}}
 \renewcommand{\algorithmicensure}{\textbf{Output: $H(x)=\sum_{t=1}^T \alpha_th_t(x)$}} 
 \REQUIRE 
 \ENSURE  
\item[]
\STATE Initialisation : $D^1(i) = 1/n$
  \FOR {$t = 1$ to $T$}
    \STATE Train weak learner $h_t\rightarrow y$ using $D^t$
    \STATE Update $\alpha_t$ based on Eq.~\eqref{eq:adaboost_a} 
    \STATE For $i = 1,...,n$:
    \STATE \ \ \ \ Update $D^{t+1}(i)$ based on Eq.~\eqref{eq:adaboost_d}
    \\ \ \ \ \  //where $Z_t$ is the normalization factor according to Eq.~\ref{eq:Z_adaboost}. 
  \ \ \ENDFOR
  \STATE return $H(x)$  // Eq.~\eqref{eq:H_adaboost}
 \end{algorithmic} 
 \end{algorithm}

Cost-sensitive boosting approaches extend  AdaBoost for class imbalance by changing the following components:
i) \emph{weight initialization} (recall that in Adaboost all instances receive the same weight during initialization - line 1 of Algorithm~\ref{alg:adaboost}), 
ii) \emph{distribution reweighting} (for AdaBoost the update is according to Eq.~\eqref{eq:adaboost_d} and Eq.~\eqref{eq:adaboost_a}), and 
iii) \emph{voting schema} (for Adaboost voting is according to Eq.~\eqref{eq:H_adaboost}). 
A detailed overview of the cost-sensitive methods and how they implement the aforementioned (i)-(iii) aspects is presented in Table~\ref{tbl:baselines}. CGAda~\cite{landesa2012shedding,landesa2015revisiting_a,landesa2015revisiting_b} employs the misclassification cost matrix only for initializing the weight distribution at the first boosting round and proceeds as standard AdaBoost thereafter. AdaCost ($\beta_2$)~\cite{fan1999adacost}, AdaC1-C3~\cite{sun2007cost} and CSB1/2~\cite{ting2000comparative} incorporate the misclassification cost matrix to change the data distribution in each boosting round. AdaMEC~\cite{nikolaou2015calibrating} and RareBoost~\cite{joshi2001evaluating} differ from the other cost-sensitive methods: In particular, AdaMEC does not use costs to change the data distribution but it rather shifts the decision boundary of AdaBoost to minimize the total expected loss. 
RareBoost does not rely on misclassification costs, rather it employs instead of a single parameter $\alpha$ (see Eq.~\eqref{eq:adaboost_a}), two different parameters, $\alpha^+$ and $\alpha^-$ for positive and negative predictions, respectively to update the weight distribution as well as the voting schema. RareBoost requires that $TP > FP$; however, if this assumption does not hold the algorithm's performance deteriorates~\cite{sun2007cost}. CGAda-Cal.~\cite{nikolaou2016cost} and AdaMEC-Cal.~\cite{nikolaou2016cost} are not shown in Table~\ref{tbl:baselines} since calibration, through Platt scaling, is applied to the trained CGAda and AdaMEC models, respectively. 

\begin{sidewaystable}\centering
\vspace{12.3cm}
\caption{An overview of cost-sensitive boosting methods w.r.t. cost assignment,initialization, distribution update and ensemble decision rule. Our methods are shown in bold. RareBoost ($^*$) uses $\alpha^+$ ($\alpha^-$) for reweighting the samples which are positively (negatively) classified.}
\label{tbl:baselines}
\Huge
\begin{adjustbox}{height=0.27\textheight}
\begin{tabular}{lccccc}
\hline
Method & Costs & Initial $D^1$ & $D^{t+1}$ update & $\alpha_{t+1}$ update & Decision H(x)\\ \hline
AdaBoost~\cite{schapire1999brief} & None & $1/n$ & $D^t(i)\exp{(-\alpha_ty_ih_t)}$ & $\frac{1}{2}\log\frac{\sum\limits_{i,h_t(x_i)=y_i}D^t(i)}{\sum\limits_{i,h_t(x_i)\neq y_i}D^t(i)}$ & $\text{sign}(\sum_{t} \alpha_th_t(x))$\\
CGAda~\cite{landesa2012shedding,landesa2015revisiting_a,landesa2015revisiting_b} & Fixed user param. & $\frac{C_i}{\sum_i^n C_i}$ & '' & '' & ''\\
AdaMEC~\cite{nikolaou2015calibrating} &'' & $1/n$ & '' & '' & $\text{sign}(\sum\limits_{y \in \{+1,-1\}}c(y)\sum\limits_{t:h_t=y} \alpha_th_t(x))$\\
RareBoost~\cite{joshi2001evaluating} & None & '' & $D^t(i)\exp{(-\alpha_t^{+,-}y_ih_t)}^*$ & $\alpha_t^{+} = \frac{1}{2}\ln{\frac{TP_t}{FP_t}}$, $\alpha_t^{-} = \frac{1}{2}\ln{\frac{TN_t}{FN_t}}$ & $\text{sign}(\sum\limits_{t:h_t(x) \geq 0} a^{+}_t h_t(x) + \sum\limits_{t:h_t(x) < 0} a^{-}_t h_t(x))$\\ 
CSB1~\cite{ting2000comparative} & Fixed user param. & $\frac{C_i}{\sum_i^n C_i}$ & $D^t(i)C_i\exp{(-y_ih_t)}$ & $\frac{1}{2}\log\frac{\sum\limits_{i,h_t(x_i)=y_i}D^t(i)}{\sum\limits_{i,h_t(x_i)\neq y_i}D^t(i)}$ & $\text{sign}(\sum_{t}\alpha_th_t(x))$\\
CSB2~\cite{ting2000comparative} & '' & '' & $D^t(i)C_i\exp{(-\alpha_ty_ih_t)}$ & '' & ''\\
AdaCost ($\beta_2$)~\cite{fan1999adacost} & '' & '' & $D^t(i)\exp{(-\alpha_ty_ih_t\beta_i)}$ & $\frac{1}{2}\log\frac{1 + \sum_{i} D^t(i)\exp{(-\alpha_ty_ih_t\beta_i)}}{1 - \sum_{i} D^t(i)\exp{(-\alpha_ty_ih_t\beta_i)}}$ & ''\\
AdaC1~\cite{sun2007cost} & '' & '' & $D^t(i)\exp{(-C_i\alpha_ty_ih_t)}$ & $\frac{1}{2}\log\frac{1 + \sum\limits_{i,y_i=h_t(x_i)}C_iD^t(i) - \sum\limits_{i,y_i \neq h_t(x_i)}C_iD^t(i)}{1 - \sum\limits_{i,y_i=h_t(x_i)}C_iD^t(i) + \sum\limits_{i,y_i \neq h_t(x_i)}C_iD^t(i)}$ & ''\\
AdaC2~\cite{sun2007cost} & '' & '' & $D^t(i)C_i\exp{(-\alpha_ty_ih_t)}$ & $\frac{1}{2}\log\frac{\sum\limits_{i,h_t(x_i)=y_i}C_iD^t(i)}{\sum\limits_{i,h_t(x_i)\neq y_i}C_iD^t(i)}$ & ''\\
AdaC3~\cite{sun2007cost} & '' & '' & $D^t(i)C_i\exp{(-C_i\alpha_ty_ih_t)}$ & $\frac{1}{2}\log\frac{\sum_iC_iD^t(i) + \sum\limits_{i,h_t(x_i)=y_i}C^2_iD^t(i) - \sum\limits_{i,h_t(x_i)\neq y_i}C^2_iD^t(i)}{\sum_iC_iD^t(i) - \sum\limits_{i,h_t(x_i)=y_i}C^2_iD^t(i) + \sum\limits_{i,h_t(x_i)\neq y_i}C^2_iD^t(i)}$ & ''\\
\textbf{AdaCC1} & Eq.~\eqref{eq:costs} & $1/n$ & $D^t(i)\exp{(-C_i^{t}\alpha_ty_ih_t)}$ & $\frac{1}{2}\log\frac{1 + \sum\limits_{i,y_i=h_t(x_i)}C^t_iD^t(i) - \sum\limits_{i,y_i \neq h_t(x_i)}C^t_iD^t(i)}{1 - \sum\limits_{i,y_i=h_t(x_i)}C^t_iD^t(i) + \sum\limits_{i,y_i \neq h_t(x_i)}C^t_iD^t(i)}$ & ''\\
\textbf{AdaCC2} & Eq.~\eqref{eq:costs} & '' & $D^t(i)C_i^{t}\exp{(-\alpha_ty_ih_t)}$ & $\frac{1}{2}\log\frac{\sum\limits_{i,h_t(x_i)=y_i}C^t_iD^t(i)}{\sum\limits_{i,h_t(x_i)\neq y_i}C^t_iD^t(i)}$ & ''\\
\hline
\end{tabular}\normalsize
\end{adjustbox}
\end{sidewaystable}

%% file: method.tex
Instead of assuming a fixed misclassification cost matrix, \ourmethod~ dynamically adjusts the misclassification costs in each boosting round based on the performance of the model up to that round, i.e., the performance of the partial ensemble (Section~\ref{sec:amortsed_cost}). 
This way, in each boosting round \ourmethod~ boosts the class with the highest misclassification rate. These costs are then used to update the data distribution for the next round. There are two ways to incorporate the costs in the update formula (for AdaBoost the update formula is shown in Eq.~\eqref{eq:adaboost_d}): inside or outside the exponent resulting in two variations AdaCC1 (Section~\ref{sec:AdaCC1}) and AdaCC2 (Section~\ref{sec:AdaCC2}), respectively. 

The toy example in Figure~\ref{fig:toy_example} demonstrates how our approach ``pays extra attention'' to the minority class errors: in particular, we observe that AdaBoost, AdaCC1 and AdaCC2 misclassify the minority class (blue points) during the first boosting round $t=1$; however, our methods assign higher weights to the minority examples on the next boosting rounds in contrast to AdaBoost, which lead to substantially different decision boundaries on the upcoming boosting rounds and also the final ensemble.

\subsection{Cumulative Misclassification Costs}
\label{sec:amortsed_cost}

Let $t \in [1,T]$ be the current boosting round, where $T$ is a user defined parameter indicating the number of boosting rounds. Let $H_{1:t}(x) = sign(\sum_{j=1}^t \alpha_jh_j(x))$ be the partial ensemble up to round $t$. 
We monitor the cumulative error of the partial ensemble and in particular, the cumulative false positive rate (FPR) and the cumulative false negative rate (FNR) defined as follows: 
\begin{equation}
\label{eq:fpr_fnr}
\begin{split}
FNR_{1:t} = \frac{\sum\limits_{i,x_i \in D_+} \mathbb{I}\left\{\text{sign}\left(\sum\limits_{j=1}^t \alpha_jh_j(x_i)\right) \neq y_i \right\}}{|D_+|} \\
FPR_{1:t} = \frac{\sum\limits_{i,x_i\in D_-}  \mathbb{I}\left\{\text{sign}\left(\sum\limits_{j=1}^t \alpha_jh_j(x_i)\right) \neq y_i\right\}}{|D_-|}
\end{split}
\end{equation}\normalsize
where $\mathbb{I}\{\cdot\}$ is the indicator function that returns 1 if the condition within is true and 0, otherwise.  The term $FNR_{1:t}$ corresponds to the error of the partial ensemble in the positive class ($D_+$); likewise,  $FPR_{1:t}$ refers to the error in the negative class ($D_-$).

Based on the cumulative error rates, we define the \emph{cumulative misclassifications costs} below in order to ``bias'' the weighting process for the next round towards the class with the highest misclassification rate (on the current boosting round):

\begin{equation}
    \label{eq:costs}
      C^{t}(x_i)=
    \begin{cases}
    1 + FNR_{1:t}, & \text{if } h_t(x_i) \neq y_i, y_i = +, FNR_{1:t} > FPR_{1:t}\\
    1 + FPR_{1:t}, & \text{if } h_t(x_i) \neq y_i, y_i = -, FNR_{1:t} < FPR_{1:t}\\
    1, & otherwise
    \end{cases}
\end{equation}\normalsize
where $h_t$ is the weak learner at round $t$. 
In particular, for any  misclassified instance $x_i$, we increase its weight using the cumulative FPR or FNR values based on its class-membership. 

The costs are therefore dynamically adjusted based on the partial ensemble's cumulative behavior and the predictions of the current weak learner. In contrast to other methods that assume \textit{fixed} misclassification costs through the boosting rounds, our method is not only \textit{parameter-free} but it also dynamically detects which class might require extra weighting at each round. We should highlight that the cumulative misclassification costs aim to boost the class with the highest misclassification rate and not individual examples. Nonetheless, the cumulative misclassification costs affect the weights of the instances since they are used to update the data distribution. 
In what follows, and when it is clear from the context, we simplify the notation of $C^t(x_i)$ as $C^t_i$.  

The two variants AdaCC1 (Section~\ref{sec:AdaCC1}) and AdaCC2 (Section~\ref{sec:AdaCC2}) are presented next.

\subsection{AdaCC1}
\label{sec:AdaCC1}
The first proposed algorithm, AdaCC1, modifies the weight update formula of AdaBoost (Eq.~\eqref{eq:adaboost_d}) using the cumulative costs $C^{t}_i$ (Eq.~\eqref{eq:costs}) as follows:
\begin{equation}
\label{eq:AdaCC1}
D^{t+1}(i) = \frac{D^t(i)\exp{\left(-C^{t}_i\alpha_t y_i h_t(x_i)\right)}}{Z_t}
\end{equation}
The normalization factor $Z_t$ (for Adaboost shown in
Eq.~\eqref{eq:Z_adaboost}), in round $t$, is also updated to take the extra weighting factor into account: 
\begin{align}
\label{eq:AdaCC1_assumptions}
Z_t = \sum\limits_{i=1}^n D^t(i)\exp{(-C^t_i\alpha_ty_ih_t(x_i))}
\end{align}\normalsize

\noindent\textbf{Error analysis}: 
By unravelling Eq.~\eqref{eq:AdaCC1}, the following holds:
\begin{equation}
\label{eq:unravel_AdaCC1}
\begin{split}
D^{t+1}(i) & = D^1(i)\times\frac{\exp{\left(-C^{1}_i\alpha_1y_ih_1(x_i)\right)}}{Z_1}\times\cdots \times\frac{\exp{\left(-C^{t}_i\alpha_ty_ih_t(x_i)\right)}}{Z_t} = \\
& = \frac{D^1(i)\exp{\left(-\sum\limits_{j=1}^tC^{j}_i \alpha_j y_ih_j(x_i) \right)}}{\prod\limits_{j=1}^t Z_j}
\end{split}
\end{equation}\normalsize

The upper bound of the training error of the final ensemble $H(x)$ can be expressed as:
\begin{equation}
\label{eq:error_AdaCC1}
Pr_{i\sim D^1} [H(x_i) \neq y_i]  \leq \sum\limits_{i=1}^n D^1(i)\exp{\left(-\sum\limits_{t=1}^T C^{t}_i \alpha_t y_ih_t(x_i) \right)} = \prod\limits_{t=1}^T Z_t
\end{equation}\normalsize
Therefore, the objective in each boosting round is to find the $\alpha_t$ that minimizes $Z_t$. Since $Z_t$ is the weight summation of correctly and non-correctly classified instances at round $t$, following the same argumentation as in~\cite{schapire1999improved,sun2007cost}, Eq.~\eqref{eq:AdaCC1_assumptions} can be expressed as: 


\begin{equation}\footnotesize
\label{eq:adacc1_at}
\begin{split}
 & \sum\limits_{i=1}^n D^t(i)\exp{\left(- C^t_i \alpha_t y_i h_t(x_i)  \right)} \leq \sum\limits_{i=1}^n D^t(i)\left( \frac{1-C^t_iy_ih_t(x_i)}{2}\exp{(\alpha_t)} + \frac{1+C^t_iy_ih_t(x_i)}{2}\exp{(-\alpha_t)}  \right)\\
    \end{split}
\end{equation}\normalsize
By differentiating Eq.~\eqref{eq:adacc1_at} w.r.t. $\alpha_t$ and setting it to zero, we can estimate $\alpha_t$ as follows:
\begin{equation}\footnotesize
\label{eq:adacc1_at_final}
\begin{split}
    & \frac{\partial }{\partial \alpha_t} \Bigg( \sum\limits_{i=1}^n D^t(i)\left( \frac{1-C^t_iy_ih_t(x_i)}{2}\exp{(\alpha_t)}\right)  + \sum\limits_{i=1}^n D^t(i)\left( \frac{1+C^t_iy_ih_t(x_i)}{2}\exp{(-\alpha_t)} \right) \Bigg)  = 0 \Rightarrow\\
    & e^\alpha_t\sum\limits_{i=1}^N D^t(i)\left( \frac{1-C^t_iy_ih_t(x_i)}{2}\right) = e^{-\alpha_t}\sum\limits_{i=1}^N D^t(i)\left( \frac{1+C^t_iy_ih_t(x_i)}{2}\right)\Rightarrow\\
     &   \alpha_t = \frac{1}{2}\log\left( \frac{\sum\limits_{i=1}^n D^t(i)(1+C^t_iy_ih_t(x_i))}{\sum\limits_{i=1}^n D^t(i)(1-C^t_iy_ih_t(x_i))} \right) = \frac{1}{2}\log\left( \frac{1 + \sum\limits_{i,y_i=h_t(x_i)}^n C_i^tD^t(i) - \sum\limits_{i,y_i\neq h_t(x_i)}^n C_i^tD^t(i)}{1 - \sum\limits_{i,y_i=h_t(x_i)}^n C_i^tD^t(i) + \sum\limits_{i,y_i\neq h_t(x_i)}^n C_i^tD^t(i)} \right)
    \end{split}
\end{equation}\normalsize
To ensure that $\alpha_t$ is non-negative, the following condition should hold, otherwise the iteration process terminates:
\begin{equation}
\label{eq:a_AdaCC1}
\sum\limits_{i,y_i=h(x_i)} C^t_iD^t(i) > \sum\limits_{i,y_i\neq h(x_i)} C^t_iD^t(i) 
\end{equation}\normalsize

\noindent\textbf{Time complexity}: We derive the time complexity of our approach building upon the complexity of AdaBoost (c.f., Algorithm~\ref{alg:adaboost}). AdaBoost complexity is $O(T\cdot(f + n))$, where $T$ is the number of boosting rounds, $O(f)$ is the complexity of a weak learner (for decision stumps it is $O(n\cdot m)$ for training and $O(n)$ for testing, where $m$ is the number of features and $n$ the number of instances~\cite{su2006fast}), and $O(n)$ is the complexity for the weight update of the instances. Our only addition to the algorithm (computationally) is the calculation of the cumulative errors (Eq.~\eqref{eq:fpr_fnr}). This computation can be reduced to $O(n)$ by maintaining a vector $\vec{o}$ of size $n$ over the boosting rounds which averages the decision outcomes of the weak learners in each boosting round. Note that the vector $\vec{o}$ is updated on each round based on the current weak learner's predictions (on the training). By doing this, we avoid spending $O(t \cdot f)$ on each boosting round $t$ i.e., we avoid the prediction time of the partial ensemble (on the training set) on each boosting round. Therefore, the complexity of AdaCC1 is: $O(T\cdot(f + 2n)) \Rightarrow O(T\cdot(f + n))$, since 2 is a constant.

\subsection{AdaCC2}
\label{sec:AdaCC2}
The second proposed algorithm, AdaCC2, modifies the weight update formula of AdaBoost (Eq.~\eqref{eq:adaboost_d}) using the cumulative costs (Eq.~\eqref{eq:costs}) as follows:

\begin{equation}
\label{eq:AdaCC2}
D^{t+1}(i) = \frac{D^t(i)C^{t}_i\exp{(-\alpha_ty_ih_t(x_i)})}{Z_t}
\end{equation}\normalsize
Similarly to AdaCC1, the normalization factor $Z_t$ is also updated to ensure  $D^{t+1}$ is still a probability distribution:
\begin{align}
\label{eq:AdaCC2_assumptions}
Z_t = \sum\limits_{i=1}^n D^t(i) C^t_i\exp{(-\alpha_ty_ih_t(x_i))}
\end{align}\normalsize

\noindent\textbf{Error analysis}: 
Following the same logic as in Eq.~\eqref{eq:AdaCC1} for AdaCC1, by unravelling Eq.~\eqref{eq:AdaCC2}, we obtain the following:
\begin{equation}
\label{eq:unravel_AdaCC2}
\begin{split}
D^{t+1}(i) = \frac{D^1(i)\prod\limits_{j=1}^t C^{j}_i\exp{(-\alpha_j y_i h_j(x_i))}}{\prod\limits_{j=1}^t Z_j} 
\end{split}
\end{equation}\normalsize
%

Similarly to AdaCC1, the upper bound of the training error of the final ensemble $H(x)$ is given by:

\begin{equation}
\label{eq:error_AdaCC2}
Pr_{i\sim D^1} [H(x_i) \neq y_i] \leq \sum\limits_{i=1}^n  D^1(i) \prod\limits_{t=1}^T C^{t}_i\exp{(-\alpha_t y_i h_t(x_i))} = \prod\limits_{t=1}^TZ_t
\end{equation}\normalsize



Following a similar to AdaCC1 rationale (Eqs.~\eqref{eq:adacc1_at} and \eqref{eq:adacc1_at_final}), the  $\alpha_t$ that minimizes $Z_t$ is given by:
\begin{equation}
\label{eq:adacc2_at_final}
     \alpha_t = \frac{1}{2}\log\left( \frac{\sum\limits_{i,y_i=h_t(x_i)}^n C_i^tD^t(i) }{\sum\limits_{i,y_i\neq h_t(x_i)}^n C_i^tD^t(i)} \right)
\end{equation}\normalsize
To ensure that $\alpha_t$ is non-negative, the same condition as in Eq.~\eqref{eq:a_AdaCC1} for AdaCC1 should hold, otherwise the iteration process terminates.

\noindent\textbf{Time complexity}: AdaCC2 has the same time complexity as AdaCC1 since their only difference pertains to the weight estimation.

%% file: evaluation_setup.tex
\rev{We compare our proposed AdaCC1 and AdaCC2 against 12 state-of-the-art cost-sensitive boosting approaches (Section~\ref{sec:baselines}) as well as 3 data level methods (SMOTE, Random Over-Sampling and Random Under-Sampling) and 2 model-based methods (SMOTEBoost and RUSBoost) using suitable class imbalance performance evaluation metrics (Section~\ref{sec:metrics}). 
We have experimented with a  large number of real-world datasets (27), depicting various characteristics in terms of class imbalance, dimensionality and cardinality. An overview of the datasets is provided in Table~\ref{tbl:datasets}. We have used the same pre-processing method on all datasets whenever categorical data were present i.e., one-hot encoding. The employed structures were numpy arrays for all datasets. In addition, all the classification methods which have been employed in this paper were trained on the exact same pre-processed data. } 
The goal of our evaluation is two-fold: to compare the different methods in terms of their predictive performance for both classes  (Section~\ref{sec:performance}), and to analyze and compare the internal behavior of our methods with the other approaches in order to understand/explain our methods' superior performance  (Section~\ref{sec:internal_behavior}).

For our experiments\footnote{Source code and data are available at: \url{https://github.com/iosifidisvasileios/CumulativeCostBoosting}}, we use decision stumps, i.e., decision trees of depth 1, as weak learners for all methods. 
Regarding the number of weak learners $T$, we experiment with different numbers $T \in [25, 50,100,200]$.
For the predictive performance experiments (Section~\ref{sec:performance}), we report on the average of 10 x 5-fold cross validation. These results are also used for the significance test of Friedman using Bonferroni correction for validating significance on multiple datasets across various methods~\cite{demvsar2006statistical}. For the experiments on the internal behavior (Section~\ref{sec:internal_behavior}), we do not perform any split rather we train on the complete datasets. By using the entire datasets for training, we avoid fluctuating values which can make the internal analysis of our methods misleading.

\input{datasets}

\subsection{Performance Metrics}
\label{sec:metrics}
\input{performance_metrics}

\subsection{Competitors and Parameter Selection}
\label{sec:baselines}
\input{baselines}

%% file: datasets.tex
\begin{table}\centering
\caption{Datasets.}
\label{tbl:datasets}
\begin{adjustbox}{width=1\columnwidth,center}
\begin{tabular}{lrrrrr}
\hline
Dataset & Features & Minority & Majority & Ratio (\textit{Min}:\textit{Maj}) & Source\\ \hline
abalone & 10 & 391 & 3,786 & 1:9.68 & \cite{Dua:2019}\\
adult census & 14 & 11,202 & 33,973 & 1:3.03 & \cite{Dua:2019}\\
bank & 16 & 4,667 & 35,337 & 1:7.57 & \cite{Dua:2019}\\
car eval. & 21 & 134 & 1,594 & 1:11.90 & \cite{Dua:2019}\\
coil 2000 & 85 & 586 & 9,236 & 1:15.76 & \cite{Dua:2019}\\
credit & 23 & 6,636 & 23,364 & 1:3.52 & \cite{Dua:2019}\\
eeg eye & 14 & 6,723 & 8,257 & 1:1.23 & \cite{Dua:2019}\\
electricity & 8 & 19,237 & 26,075 & 1:1.36 & \cite{harries1999splice}\\
isolet & 617 & 600 & 7,197 & 1:11.99 & \cite{Dua:2019}\\
letter img. & 16 & 734 & 19,266 & 1:26.25 & \cite{Dua:2019}\\
mammography & 6 & 260 & 10,923 & 1:42.01 & \cite{Dua:2019}\\
musk2 & 166 & 1,017 & 5,581 & 1:5.49 & \cite{Dua:2019}\\
optical digits & 64 & 554 & 5,066 & 1:9.14 & \cite{Dua:2019}\\
ozone level & 72 & 73 & 2,463 & 1:33.74 & \cite{Dua:2019}\\
pen digits & 16 & 1,055 & 9,937 & 1:9.42 & \cite{Dua:2019}\\
phoneme & 5 & 1,586 & 3,818 & 1:2.41 & \cite{o1984esprit}\\
protein hom. & 74 & 1,296 & 144,455 & 1:111.46 & \cite{nikolaou2016cost}\\
satimage & 36 & 626 & 5,809 & 1:9.28 & \cite{Dua:2019}\\
scene & 294 & 177 & 2,230 & 1:12.60 & \cite{nikolaou2016cost}\\
sick euthyroid & 42 & 293 & 2,870 & 1:9.80 & \cite{nikolaou2016cost}\\
skin & 3 & 50,859 & 194,198 & 1:3.82 & \cite{Dua:2019}\\
spambase & 53 & 1,813 & 2,788 & 1:1.54 & \cite{Dua:2019}\\
thyroid sick & 52 & 231 & 3,541 & 1:15.33 & \cite{nikolaou2016cost}\\
us crime & 100 & 150 & 1,844 & 1:12.29 & \cite{nikolaou2016cost}\\
webpage & 300 & 981 & 33,799 & 1:34.45 & \cite{nikolaou2016cost} \\
wilt & 5 & 261 & 4,578 & 1:17.54 & \cite{Dua:2019}\\
wine quality & 11 & 183 & 4,715 & 1:25.77 & \cite{Dua:2019}\\
\hline
\end{tabular}
\end{adjustbox}
\end{table}

%% file: performance_metrics.tex
Due to the imbalanced nature of the learning problem, we report on AUC, balanced accuracy, f1-score, gmean, TNR, and TPR. 
By following similar logic as~\cite{ditzler2012incremental}, we also use a combined overall performance measure (OPM), which averages the aforementioned metrics, since no algorithm outperforms others in all datasets and metrics. All metrics (except AUC which employs the confidence scores of the predictions) can be derived from the confusion matrix of Table~\ref{tbl:conf_matrix} as shown in Table~\ref{tbl:measures}. 

\begin{table}
\caption{Performance Metrics.}
\label{tbl:measures} 
\begin{adjustbox}{width=1\columnwidth,center}
\begin{tabular}{lc}\hline
Metric & Definition \\\hline
TPR (also Recall) & $TP/(TP+FN)$\\
TNR & $TN/(TN+FP)$\\
balanced accuracy  & $1/2 \cdot (TPR+TNR)$ \\
f1-score & $2 \cdot TP / (2 \cdot TP + FP + FN)$ \\ 
gmean & $\sqrt{TPR\cdot TNR}$ \\
OPM & $1/6 \cdot (AUC + bal.acc+ gmean + f1+ TPR + TNR)$\\
\hline
\end{tabular} 
\end{adjustbox}
\end{table}

Due to the high amount of datasets, we cannot report on each individual dataset and therefore, similarly to~\cite{ditzler2012incremental,nikolaou2016cost,yin2013empirical}, we omit individual dataset results, and report on the average across all datasets. 

%% file: baselines.tex
Our main competitors are \nrcompetitors{} cost-sensitive boosting methods, namely, AdaCost ($\beta_2$)~\cite{fan1999adacost}, AdaC1~\cite{sun2007cost}, AdaC2~\cite{sun2007cost}, AdaC3~\cite{sun2007cost}, AdaMEC~\cite{nikolaou2015calibrating}, AdaMEC-Cal.~\cite{nikolaou2016cost}, CGAda~\cite{landesa2012shedding,landesa2015revisiting_a,landesa2015revisiting_b}, CGAda-Cal.~\cite{nikolaou2016cost}, CSB1~\cite{ting2000comparative}, CSB2~\cite{ting2000comparative}, and RareBoost~\cite{joshi2001evaluating}. We also employ the vanilla AdaBoost~\cite{schapire1999brief} to show the differences between cost-sensitive and standard boosting methods. 
The methods (including ours) are summarized in terms of their key characteristics in Table~\ref{tbl:baselines} (as already mentioned, AdaMEC-Cal. and CGAda-Cal. are excluded since they are the post-processed versions of AdaMEC and CGAda, respectively). Except for the AdaBoost, RareBoost and our AdaCC1 and AdaCC2 methods, all other methods need to be initialized with the misclassification cost matrix [$C_+, C_-$]. As already discussed, finding the right costs is a tedious task requiring domain/dataset knowledge. To this end, we follow the suggestion of~\cite{sun2007cost,nikolaou2016cost} to use grid search for selecting the best class ratio for misclassification costs. In particular, for each dataset, we perform grid search on a variety of different class ratios, namely with $C_+ = 1.0$ and by varying $C_-$ in the range $[0.1 - 1.0]$ with step $0.1$. We select the class ratio which achieves the best f1-score as suggested by~\cite{sun2007cost,nikolaou2016cost}. Grid search is performed on each fold (on the training set) and each value of $T \in [25, 50, 100, 200]$; therefore, for all 10 iterations and for each different fold, the competitors are fine-tuned\footnote{Note: We have also used a validation set for tuning the competitors by splitting the training set into 80\% training 20\% validation (on each fold); however, the results were slightly worse, hence we have tuned competitors on the training set.}.

\rev{We have combined the three data-level methods  with a decision tree classifier. We augmented the minority class until the class-imbalance was eliminated i.e., both classes had the same amount of instances. For the under-sampling we also removed instances from the majority class until both classes had the same amount of instances. For the model-level methods, we have used the default parameters e.g., for SMOTEBoost we set $k=5$ and varied the number of weak learners same as before and same for RUSBoost. }

In addition, we evaluate the impact of the cumulative misclassification costs (Eq.~\eqref{eq:costs}) which allows us to dynamically adjust the costs based on the performance of the partial ensemble and is central to our approach. To this end, we compare AdaCC1 and AdaCC2 with their non-cumulative counterparts, denoted by AdaN-CC1 and AdaN-CC2, respectively. The only difference is that the non-cumulative versions do not take into consideration the cumulative error of the partial ensemble, rather rely on each individual weak learner to estimate the misclassification costs for the next round. More concretely, the partial ensemble up to round $t$, i.e., $\sum_{j=1}^t \alpha_jh_j(x)$ in Eq.~\eqref{eq:fpr_fnr}, is replaced by the corresponding weak learner in round $t$, i.e., $h_t(x)$.

%% file: experiments.tex
We split the experiments into two categories: i) predictive performance (Section~\ref{sec:performance}) and ii) internal analysis (Section~\ref{sec:internal_behavior}). \rev{In the first category, we compare the predictive performance of our methods against other cost sensitive boosting competitors using the metrics from Section~\ref{sec:metrics}. Although the aim of this work is to compare cost-sensitive boosting methods, we also highlight in Table~\ref{tbl:non_cost_sensitive_methods} the performance of data-level methods such as SMOTE~\cite{chawla2002smote} (where the number of neighbors $k=5$), Random Under-Sampling (RUS) and Random Over-Sampling (ROS) combined with decision tree classifiers. Also, we employ  boosting class-imbalance methods such as SMOTEBoost~\cite{chawla2003smoteboost} and RUSBoost~\cite{seiffert2009rusboost}}. In the second category, we compare how our method differs from the others by showing the internal behavior of each method. 

\subsection{Predictive Performance}
\label{sec:performance}
\input{exp_performance}

\subsection{Internal Analysis}
\label{sec:internal_behavior}
\input{exp_internal_behavior}

%% file: exp_performance.tex
In this section, we begin by comparing the performance of our method against the employed competitors. We continue by comparing AdaCC with its non-cumulative counterpart AdaN-CC. Note that the performance results, in terms of different evaluation metrics shown in Tables~\ref{tbl:results} and~\ref{tbl:counter_parts_comparison}, are averaged over all datasets. 
Afterwards, we report on the ranking of each method based on the datasets. Finally, we report on the statistical significance of our results.

\noindent\textbf{AdaCC vs Competitors:} We begin our analysis for the main competitors in Table~\ref{tbl:results}. AdaCC1 and AdaCC2 are the best in terms of  balanced accuracy, gmean, recall (TPR) and OPM (AdaCC1 is also best in AUC). AdaMEC-Cal. follows with a [1.27\%-1.77\%] relative decrease in OPM (it has very close difference with AdaCC2), [3.44\%-3.57\%] relative decrease in balanced accuracy and [4.78\%-5.16\%] relative decrease in gmean comparing to our best performing method (AdaCC2). The fourth performing method is CGAda-Cal. with a [1.54\%-2.39\%] relative decrease in OPM, [4.13\%-4.49\%] decrease in balanced accuracy and [5.82\%-6.48\%] relative decrease in gmean comparing to our best performing method (AdaCC2). In terms of balanced accuracy, gmean and recall, AdaCC1 and AdaCC2 have the best performance. 
A closer look to the TPR, TNR scores shows that our approaches achieve the best performance for the minority class (higher TPR), while maintaining a moderate performance for the minority class (TNR close to average).

\input{special_table}

As expected, AdaBoost, which does not tackle imbalance, achieves the highest TNR but lowest TPR. The cost-sensitive competitors are able to produce higher TPR scores than AdaBoost, but still fail to learn the minority class effectively e.g., AdaC1, AdaC2 and AdaC3 produce [73.3\%-78.25\%] balanced accuracy, [65.44\%-75.4\%]  gmean and [61.04\%-68.21\%] TPR scores which are significantly lower in contrast to our methods.

The competitive performance of AdaMEC-Cal. and CGAda-Cal. is mainly due to their high TNR and low recall. AdaMEC-Cal.'s relative difference in recall is [13.87\%-17.28\%] lower than our approaches, and for CGAda-Cal. the relative difference is [15.7\%-17.98\%] lower. RareBoost also calls for special mention as  it performs poorly on the minority class but achieves the second best TNR scores. Its outlying behavior is probably related to its strong assumption that $TP > FP$, which cannot be always ensured.
 

The obtained results indicate that the cost-sensitive boosting competitors are producing higher balanced accuracy in contrast to AdaBoost but they fail to outperform our methods as indicated by balanced accuracy, gmean, recall, and AUC metrics. In addition, some competitors such as AdaCost, AdaC2, AdaC3, CSB1, and CSB2 do not improve their performance for higher values of $T$ in contrast to other competitors. One possible reason for the sub-optimal performance of the competitors might be the non-optimal misclassification cost tuning as a result of the grid search. Our methods avoid this by dynamically adjusting misclassification costs on each boosting round based on the cumulative behavior of the model. 

\noindent\textbf{Cumulative vs Non-Cumulative:} We continue by comparing our methods, AdaCC1 and AdaCC2, with their non-cumulative counterparts, namely AdaN-CC1 and AdaN-CC2, in Table~\ref{tbl:counter_parts_comparison}. By comparing AdaCC1 to AdaN-CC1 we observe a relative decrease of [16\% - 17.47\%] in balanced accuracy, [42.29\% - 56.36\%] in gmean, [37.79\% - 63.47\%] and [10.01\% - 19.94\%] in AUC. There are also high (relative) differences between AdaCC2 and AdaN-CC2. These differences highlight the superiority of the cumulative costs in the reweighting procedure on each boosting round versus the non-cumulative costs.

\noindent\textbf{Ranking:} We also report on the ranks based on balanced accuracy across the methods in Table~\ref{tbl:ranks}, for $T=200$ (Tables for $T \in [25, 50, 100]$ are included in the Appendix). Note that Table~\ref{tbl:ranks} contains floats instead of integers due to the fact that in many datasets some methods produced the same balanced accuracy score.

There are some interesting observations from this table. AdaCC1 and AdaCC2 are the best and second-best in ranks with an average rank of 2.06 and 3.19 respectively, in contrast to the competitors; however, methods such as AdaMEC-Cal. and CGAda-Cal. are also achieving high ranks. Furthermore, the last row of Table~\ref{tbl:ranks} shows the number of datasets for which each method achieved the best performance. Our approaches, AdaCC1 and AdaCC2, have won on 10 and 8 datasets, while for the majority of datasets AdaCC1 or AdaCC2 were the best or second best methods. Similar behavior can also be observed for other values of $T$, where AdaCC1 and AdaCC2 achieve the best ranking scores e.g., AdaCC1 achieves the best ranking for $T \in [25,50,100]$ with values 2.30, 2.33 and 2.11, respectively and AdaCC2 achieves the second best ranking with scores 2.70, 2.41 and 2.52.

\rev{In Table~\ref{tbl:non_cost_sensitive_methods}, we also compare non cost-sensitive methods with our approach. We have used three well-known data-level methods such as SMOTE, Random Over-Sampling (ROS) and Random Under-Sampling (RUS) combined with a decision tree classifier, and also two model-based boosting methods such as SMOTEBoost and RUSBoost. As we can see, AdaCC performs better than the other methods in terms of balanced accuracy, gmean, auc and OPM. It is also visible that RUSBoost is able to maintain extremely high TPR scores; however, it under-performs in terms of TNR in contrast to AdaCC which maintains both TNR and TPR at high levels. Interestingly, by comparing Table~\ref{tbl:results} and Table~\ref{tbl:non_cost_sensitive_methods}, we can observe that the non cost-sensitive methods are able to outperform several cost-sensitive methods.}

\noindent\textbf{Statistical Significance:} Finally, for the comparison of cost-sensitive methods we have performed the Friedman test ($p < 0.05$) using the Bonferroni correction~\cite{demvsar2006statistical} for comparing multiple methods across multiple datasets. The results can be seen in Table~\ref{tbl:sig_test}, in which non-significant values have been highlighted in bold. As we see, AdaCC1 and AdaCC2 are not significantly different across various values of $T$. AdaCC1 and AdaCC2 are significantly different compared to the other competitors. One interesting observation is that for high $T$, AdaMEC-Cal. and CGAda-Cal. are able to produce similar results as our methods.

\begin{table}[tp!]\centering
\caption{Results for various evaluation metrics for non-cost-sensitive class-imbalance methods. Best and second best methods per different values of $T$ are in bold and circled, respectively. Colors indicate specific values of T.}
\label{tbl:non_cost_sensitive_methods} 
\begin{adjustbox}{width=1\textwidth}
\begin{tabular}{ccccccccc}
\hline
Method & T & Bal. Acc & Gmean & TPR (Recall) & TNR & F1Score & AUC & OPM \\ \hline
\multirow{4}{*}{AdaCC1}      
 & \cellcolor{Gray0} 25 & \cellcolor{Gray0} \textbf{83.16$\pm$3.29} & \cellcolor{Gray0} \circletext{82.11$\pm$6.23} & \cellcolor{Gray0} \circletext{79.68$\pm$8.42} & \cellcolor{Gray0} 86.65$\pm$3.65 & \cellcolor{Gray0} 55.8$\pm$4.86 & \cellcolor{Gray0} \textbf{90.28$\pm$1.83} & \cellcolor{Gray0} \textbf{79.61$\pm$3.57} \\ 
 & \cellcolor{Gray1} 50 & \cellcolor{Gray1} \textbf{84.25$\pm$2.64} & \cellcolor{Gray1} \circletext{83.53$\pm$4.56} & \cellcolor{Gray1} \circletext{81.92$\pm$4.71} & \cellcolor{Gray1} 86.58$\pm$4.83 & \cellcolor{Gray1} 58.16$\pm$3.51 & \cellcolor{Gray1} \textbf{90.98$\pm$1.6} & \cellcolor{Gray1} \textbf{80.9$\pm$2.74} \\ 
 & \cellcolor{Gray2} 100 & \cellcolor{Gray2} \textbf{85.01$\pm$2.07} & \cellcolor{Gray2} \textbf{84.6$\pm$2.8} & \cellcolor{Gray2} \circletext{81.67$\pm$3.75} & \cellcolor{Gray2} 88.35$\pm$2.2 & \cellcolor{Gray2} 60.48$\pm$2.73 & \cellcolor{Gray2} \textbf{91.49$\pm$1.48} & \cellcolor{Gray2} \textbf{81.93$\pm$2.02} \\ 
 & \cellcolor{Gray3} 200 & \cellcolor{Gray3} \textbf{85.21$\pm$1.85} & \cellcolor{Gray3} \textbf{84.64$\pm$2.41} & \cellcolor{Gray3} \circletext{81.19$\pm$3.52} & \cellcolor{Gray3} 89.23$\pm$1.55 & \cellcolor{Gray3} 61.91$\pm$2.8 & \cellcolor{Gray3} \textbf{91.78$\pm$1.38} & \cellcolor{Gray3} \textbf{82.32$\pm$1.83} \\ \hline
\multirow{4}{*}{AdaCC2}      
 & \cellcolor{Gray0}  25 & \cellcolor{Gray0} \circletext{82.97$\pm$2.06} & \cellcolor{Gray0} \textbf{82.47$\pm$2.38} & \cellcolor{Gray0} \textbf{80.69$\pm$5.79} & \cellcolor{Gray0} 85.25$\pm$3.53 & \cellcolor{Gray0} 56.24$\pm$3.04 & \cellcolor{Gray0} \circletext{89.76$\pm$1.91} & \cellcolor{Gray0} \circletext{79.55$\pm$1.93} \\ 
 & \cellcolor{Gray1} 50 & \cellcolor{Gray1} \circletext{84.17$\pm$1.98} & \cellcolor{Gray1} \textbf{83.66$\pm$2.33} & \cellcolor{Gray1} 80.69$\pm$5.25 & \cellcolor{Gray1} 87.65$\pm$2.85 & \cellcolor{Gray1} 58.9$\pm$3.0 & \cellcolor{Gray1} \circletext{89.72$\pm$2.2} & \cellcolor{Gray1} \circletext{80.79$\pm$1.96} \\ 
 & \cellcolor{Gray2} 100 & \cellcolor{Gray2} \circletext{84.46$\pm$1.95} & \cellcolor{Gray2} \circletext{83.71$\pm$2.42} & \cellcolor{Gray2} 80.54$\pm$5.04 & \cellcolor{Gray2} 88.38$\pm$2.82 & \cellcolor{Gray2} 60.08$\pm$3.16 & \cellcolor{Gray2} 89.16$\pm$2.58 & \cellcolor{Gray2} \circletext{81.05$\pm$1.99} \\ 
 & \cellcolor{Gray3} 200 & \cellcolor{Gray3} \circletext{84.41$\pm$1.85} & \cellcolor{Gray3} \circletext{83.31$\pm$2.37} & \cellcolor{Gray3} 79.22$\pm$4.76 & \cellcolor{Gray3} \textbf{89.6$\pm$2.36} & \cellcolor{Gray3} 62.4$\pm$3.01 & \cellcolor{Gray3} 87.67$\pm$3.56 & \cellcolor{Gray3} \circletext{81.1$\pm$2.0} \\ \hline

SMOTE + D.T. &  1 & 79.41$\pm$2.11 & 76.53$\pm$2.93 & 65.24$\pm$4.17 & 93.58$\pm$0.7 & 61.31$\pm$2.91 & 79.35$\pm$2.14 & 75.9$\pm$2.34 \\ \hline
ROS + D.T. &  1 & 78.27$\pm$1.97 & 74.24$\pm$3.06 & 61.67$\pm$3.9 & \circletext{94.87$\pm$0.59} & \circletext{61.99$\pm$3.06} & 78.09$\pm$2.0 & 74.86$\pm$2.3 \\  \hline
RUS + D.T. &  1 & 82.35$\pm$1.99 & 82.29$\pm$2.0 & 82.8$\pm$3.73 & 81.9$\pm$2.12 & 50.44$\pm$2.82 & 82.34$\pm$2.01 & 77.02$\pm$2.05 \\  \hline

\multirow{4}{*}{RUSBoost}      
& \cellcolor{Gray0} 25 & \cellcolor{Gray0} 79.36$\pm$2.29 & \cellcolor{Gray0} 77.07$\pm$3.03 & \cellcolor{Gray0} 79.38$\pm$3.6 & \cellcolor{Gray0} 79.34$\pm$3.72 & \cellcolor{Gray0} 51.57$\pm$3.18 & \cellcolor{Gray0} 88.41$\pm$2.06 & \cellcolor{Gray0} 75.86$\pm$2.33 \\
& \cellcolor{Gray1} 50 & \cellcolor{Gray1} 76.35$\pm$3.02 & \cellcolor{Gray1} 71.98$\pm$4.88 & \cellcolor{Gray1} \textbf{88.92$\pm$2.29} & \cellcolor{Gray1} 63.79$\pm$5.93 & \cellcolor{Gray1} 43.98$\pm$2.93 & \cellcolor{Gray1} 86.99$\pm$2.86 & \cellcolor{Gray1} 72.0$\pm$3.05 \\
& \cellcolor{Gray2} 100 & \cellcolor{Gray2} 70.16$\pm$3.66 & \cellcolor{Gray2} 60.31$\pm$7.52 & \cellcolor{Gray2} \textbf{94.67$\pm$1.7} & \cellcolor{Gray2} 45.64$\pm$7.42 & \cellcolor{Gray2} 36.22$\pm$2.39 & \cellcolor{Gray2} 84.34$\pm$3.83 & \cellcolor{Gray2} 65.22$\pm$3.79 \\ 
& \cellcolor{Gray3} 200 & \cellcolor{Gray3} 64.61$\pm$3.94 & \cellcolor{Gray3} 48.61$\pm$9.69 & \cellcolor{Gray3} \textbf{97.8$\pm$1.21} & \cellcolor{Gray3} 31.43$\pm$8.26 & \cellcolor{Gray3} 30.29$\pm$1.96 & \cellcolor{Gray3} 81.68$\pm$4.93 & \cellcolor{Gray3} 59.07$\pm$4.34 \\ \hline
\multirow{4}{*}{SMOTEBoost}      
& \cellcolor{Gray0}  25 & \cellcolor{Gray0} 80.46$\pm$2.1 & \cellcolor{Gray0} 77.93$\pm$2.8 & \cellcolor{Gray0} 68.51$\pm$4.37 & \cellcolor{Gray0} \textbf{92.42$\pm$0.89} & \cellcolor{Gray0} \textbf{63.15$\pm$2.85} & \cellcolor{Gray0} 89.3$\pm$1.42 & \cellcolor{Gray0} 78.62$\pm$2.07 \\ 
& \cellcolor{Gray1} 50 & \cellcolor{Gray1} 81.69$\pm$2.0 & \cellcolor{Gray1} 79.49$\pm$2.61 & \cellcolor{Gray1} 71.93$\pm$4.07 & \cellcolor{Gray1} \textbf{91.45$\pm$0.85} & \cellcolor{Gray1} \textbf{64.71$\pm$2.63} & \cellcolor{Gray1} 89.71$\pm$1.39 & \cellcolor{Gray1} 79.83$\pm$1.96 \\ 
& \cellcolor{Gray2} 100 & \cellcolor{Gray2} 82.06$\pm$1.89 & \cellcolor{Gray2} 79.89$\pm$2.47 & \cellcolor{Gray2} 73.72$\pm$3.82 & \cellcolor{Gray2} \textbf{90.4$\pm$0.82} & \cellcolor{Gray2} \textbf{65.14$\pm$2.48} & \cellcolor{Gray2} \circletext{89.63$\pm$1.54} & \cellcolor{Gray2} 80.14$\pm$1.88 \\ 
& \cellcolor{Gray3} 200 & \cellcolor{Gray3}81.91$\pm$1.82 & \cellcolor{Gray3}79.58$\pm$2.43 & \cellcolor{Gray3}74.46$\pm$3.72 & \cellcolor{Gray3}89.37$\pm$0.81 &\cellcolor{Gray3} \textbf{65.14$\pm$2.33} &\cellcolor{Gray3} \circletext{89.63$\pm$1.43} &\cellcolor{Gray3} 80.01$\pm$1.79 \\ \hline

\end{tabular}\end{adjustbox}
\end{table}


%% file: special_table.tex
\begin{table}[tp!]\centering
\caption{Results for various evaluation metrics. Best and second best methods per different values of $T$ are in bold and circled, respectively. Colors indicate specific values of T.}
\label{tbl:results} 
\begin{adjustbox}{width=1\textwidth}
\begin{tabular}{ccccccccc}
\hline
Method                       & T & Bal. Acc & Gmean & TPR (Recall) & TNR & F1Score & AUC & OPM \\ \hline
\multirow{4}{*}{AdaBoost}   
 & \cellcolor{Gray0} 25 & \cellcolor{Gray0} 70.24$\pm$2.69 & \cellcolor{Gray0} 56.86$\pm$5.9 & \cellcolor{Gray0} 43.49$\pm$5.81 & \cellcolor{Gray0} \textbf{97.0$\pm$0.74} & \cellcolor{Gray0} 49.62$\pm$5.23 & \cellcolor{Gray0} 89.86$\pm$1.24 & \cellcolor{Gray0} 67.85$\pm$3.25 \\ 
 & \cellcolor{Gray1} 50 & \cellcolor{Gray1} 73.06$\pm$2.35 & \cellcolor{Gray1} 63.03$\pm$4.62 & \cellcolor{Gray1} 49.19$\pm$4.93 & \cellcolor{Gray1} \textbf{96.93$\pm$0.52} & \cellcolor{Gray1} 55.46$\pm$4.32 & \cellcolor{Gray1} \circletext{90.49$\pm$1.21} & \cellcolor{Gray1} 71.36$\pm$2.72 \\ 
 & \cellcolor{Gray2} 100 & \cellcolor{Gray2} 75.05$\pm$1.79 & \cellcolor{Gray2} 66.24$\pm$3.05 & \cellcolor{Gray2} 53.17$\pm$3.69 & \cellcolor{Gray2} \textbf{96.93$\pm$0.43} & \cellcolor{Gray2} 58.69$\pm$3.09 & \cellcolor{Gray2} \circletext{90.87$\pm$1.19} & \cellcolor{Gray2} 73.49$\pm$1.98 \\ 
 & \cellcolor{Gray3} 200 & \cellcolor{Gray3} 76.09$\pm$1.75 & \cellcolor{Gray3} 67.75$\pm$3.1 & \cellcolor{Gray3} 55.2$\pm$3.54 & \cellcolor{Gray3} \textbf{96.98$\pm$0.39} & \cellcolor{Gray3} 60.34$\pm$2.98 & \cellcolor{Gray3} \circletext{91.09$\pm$1.2} & \cellcolor{Gray3} 74.57$\pm$1.93 \\ \hline
\multirow{4}{*}{AdaCC1}      
 & \cellcolor{Gray0}  25 & \cellcolor{Gray0} \textbf{83.16$\pm$3.29} & \cellcolor{Gray0} \circletext{82.11$\pm$6.23} & \cellcolor{Gray0} \circletext{79.68$\pm$8.42} & \cellcolor{Gray0} 86.65$\pm$3.65 & \cellcolor{Gray0} 55.8$\pm$4.86 & \cellcolor{Gray0} \textbf{90.28$\pm$1.83} & \cellcolor{Gray0} \textbf{79.61$\pm$3.57} \\ 
 & \cellcolor{Gray1}  50 & \cellcolor{Gray1} \textbf{84.25$\pm$2.64} & \cellcolor{Gray1} \circletext{83.53$\pm$4.56} & \cellcolor{Gray1} \textbf{81.92$\pm$4.71} & \cellcolor{Gray1} 86.58$\pm$4.83 & \cellcolor{Gray1} 58.16$\pm$3.51 & \cellcolor{Gray1} \textbf{90.98$\pm$1.6} & \cellcolor{Gray1} \textbf{80.9$\pm$2.74} \\ 
 & \cellcolor{Gray2} 100 & \cellcolor{Gray2} \textbf{85.01$\pm$2.07} & \cellcolor{Gray2} \textbf{84.6$\pm$2.8} & \cellcolor{Gray2} \textbf{81.67$\pm$3.75} & \cellcolor{Gray2} 88.35$\pm$2.2 & \cellcolor{Gray2} 60.48$\pm$2.73 & \cellcolor{Gray2} \textbf{91.49$\pm$1.48} & \cellcolor{Gray2} \textbf{81.93$\pm$2.02} \\ 
 & \cellcolor{Gray3}  200 & \cellcolor{Gray3} \textbf{85.21$\pm$1.85} & \cellcolor{Gray3} \textbf{84.64$\pm$2.41} & \cellcolor{Gray3} \textbf{81.19$\pm$3.52} & \cellcolor{Gray3} 89.23$\pm$1.55 & \cellcolor{Gray3} 61.91$\pm$2.8 & \cellcolor{Gray3} \textbf{91.78$\pm$1.38} & \cellcolor{Gray3} \textbf{82.32$\pm$1.83} \\ \hline
\multirow{4}{*}{AdaCC2}      
 & \cellcolor{Gray0}  25 & \cellcolor{Gray0} \circletext{82.97$\pm$2.06} & \cellcolor{Gray0} \textbf{82.47$\pm$2.38} & \cellcolor{Gray0} \textbf{80.69$\pm$5.79} & \cellcolor{Gray0} 85.25$\pm$3.53 & \cellcolor{Gray0} 56.24$\pm$3.04 & \cellcolor{Gray0} 89.76$\pm$1.91 & \cellcolor{Gray0} \circletext{79.55$\pm$1.93} \\ 
 & \cellcolor{Gray1} 50 & \cellcolor{Gray1} \circletext{84.17$\pm$1.98} & \cellcolor{Gray1} \textbf{83.66$\pm$2.33} & \cellcolor{Gray1} \circletext{80.69$\pm$5.25} & \cellcolor{Gray1} 87.65$\pm$2.85 & \cellcolor{Gray1} 58.9$\pm$3.0 & \cellcolor{Gray1} 89.72$\pm$2.2 & \cellcolor{Gray1} \circletext{80.79$\pm$1.96} \\ 
 & \cellcolor{Gray2}  100 & \cellcolor{Gray2} \circletext{84.46$\pm$1.95} & \cellcolor{Gray2} \circletext{83.71$\pm$2.42} & \cellcolor{Gray2} \circletext{80.54$\pm$5.04} & \cellcolor{Gray2} 88.38$\pm$2.82 & \cellcolor{Gray2} 60.08$\pm$3.16 & \cellcolor{Gray2} 89.16$\pm$2.58 & \cellcolor{Gray2} \circletext{81.05$\pm$1.99} \\ 
 & \cellcolor{Gray3}   200 & \cellcolor{Gray3} \circletext{84.41$\pm$1.85} & \cellcolor{Gray3} \circletext{83.31$\pm$2.37} & \cellcolor{Gray3} \circletext{79.22$\pm$4.76} & \cellcolor{Gray3} 89.6$\pm$2.36 & \cellcolor{Gray3} 62.4$\pm$3.01 & \cellcolor{Gray3} 87.67$\pm$3.56 & \cellcolor{Gray3} \circletext{81.1$\pm$2.0} \\ \hline
\multirow{4}{*}{AdaMEC}      
 & \cellcolor{Gray0}   25 & \cellcolor{Gray0} 79.65$\pm$2.89 & \cellcolor{Gray0} 77.71$\pm$4.03 & \cellcolor{Gray0} 69.35$\pm$7.33 & \cellcolor{Gray0} 89.95$\pm$2.4 & \cellcolor{Gray0} 61.54$\pm$3.13 & \cellcolor{Gray0} 89.08$\pm$1.19 & \cellcolor{Gray0} 77.88$\pm$2.58 \\ 
 & \cellcolor{Gray1}  50 & \cellcolor{Gray1} 79.86$\pm$3.03 & \cellcolor{Gray1} 77.61$\pm$4.11 & \cellcolor{Gray1} 69.09$\pm$7.17 & \cellcolor{Gray1} 90.63$\pm$1.87 & \cellcolor{Gray1} 62.74$\pm$3.21 & \cellcolor{Gray1} 89.91$\pm$1.13 & \cellcolor{Gray1} 78.4$\pm$2.69 \\ 
 & \cellcolor{Gray2}  100 & \cellcolor{Gray2} 79.34$\pm$2.56 & \cellcolor{Gray2} 76.87$\pm$3.52 & \cellcolor{Gray2} 67.91$\pm$5.78 & \cellcolor{Gray2} 90.77$\pm$1.58 & \cellcolor{Gray2} 64.5$\pm$3.03 & \cellcolor{Gray2} 90.41$\pm$1.08 & \cellcolor{Gray2} 78.4$\pm$2.39 \\ 
 & \cellcolor{Gray3}  200 & \cellcolor{Gray3} 80.24$\pm$2.26 & \cellcolor{Gray3} 77.38$\pm$3.52 & \cellcolor{Gray3} 67.53$\pm$5.72 & \cellcolor{Gray3} 92.95$\pm$2.17 & \cellcolor{Gray3} 65.12$\pm$2.92 & \cellcolor{Gray3} 90.74$\pm$1.08 & \cellcolor{Gray3} 79.01$\pm$2.22 \\ \hline
\multirow{4}{*}{AdaMEC-Cal.} 
 & \cellcolor{Gray0}  25 & \cellcolor{Gray0} 80.29$\pm$2.56 & \cellcolor{Gray0} 78.44$\pm$3.5 & \cellcolor{Gray0} 68.8$\pm$6.03 & \cellcolor{Gray0} 91.78$\pm$1.56 & \cellcolor{Gray0} \circletext{62.46$\pm$3.01} & \cellcolor{Gray0} 89.9$\pm$1.33 & \cellcolor{Gray0} 78.61$\pm$2.36 \\ 
 & \cellcolor{Gray1}  50 & \cellcolor{Gray1} 81.45$\pm$2.43 & \cellcolor{Gray1} 79.84$\pm$3.23 & \cellcolor{Gray1} 70.28$\pm$5.58 & \cellcolor{Gray1} 92.62$\pm$1.35 & \cellcolor{Gray1} \circletext{64.52$\pm$2.92} & \cellcolor{Gray1} 90.36$\pm$1.4 & \cellcolor{Gray1} 79.84$\pm$2.25 \\ 
 & \cellcolor{Gray2} 100 & \cellcolor{Gray2} 82.15$\pm$2.34 & \cellcolor{Gray2} 80.48$\pm$3.17 & \cellcolor{Gray2} 71.27$\pm$5.32 & \cellcolor{Gray2} 93.03$\pm$1.24 & \cellcolor{Gray2} 65.76$\pm$2.89 & \cellcolor{Gray2} 90.3$\pm$1.41 & \cellcolor{Gray2} 80.5$\pm$2.21 \\ 
 & \cellcolor{Gray3}  200 & \cellcolor{Gray3} 82.37$\pm$2.19 & \cellcolor{Gray3} 80.48$\pm$3.07 & \cellcolor{Gray3} 71.3$\pm$4.98 & \cellcolor{Gray3} 93.44$\pm$1.11 & \cellcolor{Gray3} 66.65$\pm$2.74 & \cellcolor{Gray3} 89.72$\pm$1.47 & \cellcolor{Gray3} 80.66$\pm$2.12 \\ \hline
\multirow{4}{*}{CGAda}     
 & \cellcolor{Gray0} 25 & \cellcolor{Gray0} 79.57$\pm$2.78 & \cellcolor{Gray0} 77.45$\pm$3.9 & \cellcolor{Gray0} 67.93$\pm$6.48 & \cellcolor{Gray0} 91.21$\pm$1.77 & \cellcolor{Gray0} 62.39$\pm$3.07 & \cellcolor{Gray0} 89.83$\pm$1.29 & \cellcolor{Gray0} 78.06$\pm$2.56 \\ 
 & \cellcolor{Gray1}   50 & \cellcolor{Gray1} 80.53$\pm$2.41 & \cellcolor{Gray1} 78.58$\pm$3.35 & \cellcolor{Gray1} 68.96$\pm$5.7 & \cellcolor{Gray1} 92.09$\pm$1.61 & \cellcolor{Gray1} 64.39$\pm$2.93 & \cellcolor{Gray1} 90.42$\pm$1.22 & \cellcolor{Gray1} 79.16$\pm$2.26 \\ 
 & \cellcolor{Gray2}  100 & \cellcolor{Gray2} 81.26$\pm$2.36 & \cellcolor{Gray2} 79.36$\pm$3.26 & \cellcolor{Gray2} 69.99$\pm$5.41 & \cellcolor{Gray2} 92.53$\pm$1.31 & \cellcolor{Gray2} \circletext{65.89$\pm$2.9} & \cellcolor{Gray2} 90.79$\pm$1.19 & \cellcolor{Gray2} 79.97$\pm$2.23 \\ 
 & \cellcolor{Gray3} 200 & \cellcolor{Gray3} 81.44$\pm$2.37 & \cellcolor{Gray3} 79.27$\pm$3.38 & \cellcolor{Gray3} 69.81$\pm$5.4 & \cellcolor{Gray3} 93.08$\pm$1.21 & \cellcolor{Gray3} \circletext{66.95$\pm$2.8} & \cellcolor{Gray3} 90.99$\pm$1.24 & \cellcolor{Gray3} 80.26$\pm$2.24 \\ \hline
\multirow{4}{*}{CGAda-Cal.} 
 & \cellcolor{Gray0}  25 & \cellcolor{Gray0} 79.86$\pm$2.61 & \cellcolor{Gray0} 77.93$\pm$3.66 & \cellcolor{Gray0} 68.58$\pm$6.11 & \cellcolor{Gray0} 91.14$\pm$1.68 & \cellcolor{Gray0} \textbf{62.97$\pm$2.94} & \cellcolor{Gray0} \circletext{89.9$\pm$1.33} & \cellcolor{Gray0} 78.4$\pm$2.41 \\ 
 & \cellcolor{Gray1} 50 & \cellcolor{Gray1} 80.82$\pm$2.46 & \cellcolor{Gray1} 79.0$\pm$3.42 & \cellcolor{Gray1} 69.43$\pm$5.74 & \cellcolor{Gray1} 92.21$\pm$1.54 & \cellcolor{Gray1} \textbf{64.94$\pm$2.94} & \cellcolor{Gray1} 90.36$\pm$1.4 & \cellcolor{Gray1} 79.46$\pm$2.32 \\ 
 & \cellcolor{Gray2}  100 & \cellcolor{Gray2} 81.35$\pm$2.36 & \cellcolor{Gray2} 79.45$\pm$3.31 & \cellcolor{Gray2} 70.02$\pm$5.44 & \cellcolor{Gray2} 92.69$\pm$1.37 & \cellcolor{Gray2} \textbf{66.27$\pm$2.93} & \cellcolor{Gray2} 90.3$\pm$1.41 & \cellcolor{Gray2} 80.01$\pm$2.28 \\ 
 & \cellcolor{Gray3} 200 & \cellcolor{Gray3} 81.65$\pm$2.18 & \cellcolor{Gray3} 79.54$\pm$3.14 & \cellcolor{Gray3} 70.17$\pm$5.03 & \cellcolor{Gray3} 93.14$\pm$1.22 & \cellcolor{Gray3} \textbf{67.27$\pm$2.76} & \cellcolor{Gray3} 89.72$\pm$1.47 & \cellcolor{Gray3} 80.25$\pm$2.14 \\ \hline
\multirow{4}{*}{AdaCost}    
 & \cellcolor{Gray0}   25 & \cellcolor{Gray0} 75.72$\pm$2.75 & \cellcolor{Gray0} 71.97$\pm$4.2 & \cellcolor{Gray0} 69.5$\pm$6.24 & \cellcolor{Gray0} 81.93$\pm$4.13 & \cellcolor{Gray0} 50.43$\pm$3.53 & \cellcolor{Gray0} 83.75$\pm$2.64 & \cellcolor{Gray0} 72.22$\pm$2.81 \\ 
 & \cellcolor{Gray1}  50 & \cellcolor{Gray1} 76.43$\pm$2.74 & \cellcolor{Gray1} 73.87$\pm$4.42 & \cellcolor{Gray1} 69.62$\pm$6.87 & \cellcolor{Gray1} 83.23$\pm$5.19 & \cellcolor{Gray1} 50.7$\pm$3.52 & \cellcolor{Gray1} 84.3$\pm$2.5 & \cellcolor{Gray1} 73.03$\pm$2.83 \\ 
 & \cellcolor{Gray2} 100 & \cellcolor{Gray2} 74.72$\pm$2.71 & \cellcolor{Gray2} 70.69$\pm$4.34 & \cellcolor{Gray2} 71.99$\pm$7.01 & \cellcolor{Gray2} 77.45$\pm$4.75 & \cellcolor{Gray2} 46.25$\pm$3.35 & \cellcolor{Gray2} 83.11$\pm$2.77 & \cellcolor{Gray2} 70.7$\pm$2.69 \\ 
 & \cellcolor{Gray3} 200 & \cellcolor{Gray3} 75.15$\pm$2.72 & \cellcolor{Gray3} 70.63$\pm$4.28 & \cellcolor{Gray3} 74.91$\pm$6.63 & \cellcolor{Gray3} 75.4$\pm$5.06 & \cellcolor{Gray3} 45.11$\pm$3.26 & \cellcolor{Gray3} 81.97$\pm$2.4 & \cellcolor{Gray3} 70.53$\pm$2.74 \\ \hline
\multirow{4}{*}{CSB1}   
 & \cellcolor{Gray0}  25 & \cellcolor{Gray0} 78.07$\pm$3.89 & \cellcolor{Gray0} 74.66$\pm$6.09 & \cellcolor{Gray0} 69.21$\pm$10.25 & \cellcolor{Gray0} 86.92$\pm$4.91 & \cellcolor{Gray0} 59.78$\pm$3.95 & \cellcolor{Gray0} 88.29$\pm$1.93 & \cellcolor{Gray0} 76.16$\pm$3.5 \\ 
 & \cellcolor{Gray1} 50 & \cellcolor{Gray1} 77.8$\pm$4.68 & \cellcolor{Gray1} 73.71$\pm$7.52 & \cellcolor{Gray1} 71.52$\pm$12.18 & \cellcolor{Gray1} 84.07$\pm$6.65 & \cellcolor{Gray1} 59.01$\pm$4.75 & \cellcolor{Gray1} 88.05$\pm$2.41 & \cellcolor{Gray1} 75.69$\pm$4.2 \\ 
 & \cellcolor{Gray2}  100 & \cellcolor{Gray2} 74.25$\pm$6.07 & \cellcolor{Gray2} 67.5$\pm$11.48 & \cellcolor{Gray2} 71.8$\pm$14.83 & \cellcolor{Gray2} 76.69$\pm$11.98 & \cellcolor{Gray2} 52.34$\pm$7.3 & \cellcolor{Gray2} 86.69$\pm$4.12 & \cellcolor{Gray2} 71.54$\pm$5.96 \\ 
 & \cellcolor{Gray3}  200 & \cellcolor{Gray3} 70.17$\pm$7.57 & \cellcolor{Gray3} 59.92$\pm$16.35 & \cellcolor{Gray3} 75.6$\pm$16.68 & \cellcolor{Gray3} 64.75$\pm$19.14 & \cellcolor{Gray3} 42.86$\pm$9.69 & \cellcolor{Gray3} 84.46$\pm$6.53 & \cellcolor{Gray3} 66.29$\pm$7.76 \\ \hline
\multirow{4}{*}{CSB2}       
 & \cellcolor{Gray0}  25 & \cellcolor{Gray0} 78.21$\pm$3.62 & \cellcolor{Gray0} 75.2$\pm$5.55 & \cellcolor{Gray0} 67.4$\pm$9.16 & \cellcolor{Gray0} 89.02$\pm$3.93 & \cellcolor{Gray0} 57.69$\pm$3.34 & \cellcolor{Gray0} 89.47$\pm$1.59 & \cellcolor{Gray0} 76.17$\pm$3.24 \\ 
 & \cellcolor{Gray1}  50 & \cellcolor{Gray1} 73.97$\pm$2.52 & \cellcolor{Gray1} 66.2$\pm$4.99 & \cellcolor{Gray1} 59.17$\pm$7.38 & \cellcolor{Gray1} 88.77$\pm$4.14 & \cellcolor{Gray1} 57.11$\pm$3.83 & \cellcolor{Gray1} 90.35$\pm$1.43 & \cellcolor{Gray1} 72.6$\pm$2.74 \\ 
 & \cellcolor{Gray2}   100 & \cellcolor{Gray2} 75.07$\pm$1.86 & \cellcolor{Gray2} 66.27$\pm$3.28 & \cellcolor{Gray2} 60.57$\pm$3.69 & \cellcolor{Gray2} 89.56$\pm$0.6 & \cellcolor{Gray2} 59.74$\pm$3.1 & \cellcolor{Gray2} 90.07$\pm$1.77 & \cellcolor{Gray2} 73.55$\pm$2.11 \\ 
 & \cellcolor{Gray3} 200 & \cellcolor{Gray3} 76.09$\pm$1.73 & \cellcolor{Gray3} 67.65$\pm$2.89 & \cellcolor{Gray3} 62.6$\pm$3.52 & \cellcolor{Gray3} 89.57$\pm$0.38 & \cellcolor{Gray3} 61.37$\pm$2.94 & \cellcolor{Gray3} 90.27$\pm$1.77 & \cellcolor{Gray3} 74.59$\pm$1.99 \\ \hline
\multirow{4}{*}{AdaC1}     
 & \cellcolor{Gray0} 25 & \cellcolor{Gray0} 73.99$\pm$2.46 & \cellcolor{Gray0} 69.39$\pm$4.43 & \cellcolor{Gray0} 61.09$\pm$6.49 & \cellcolor{Gray0} 86.9$\pm$3.58 & \cellcolor{Gray0} 52.34$\pm$3.8 & \cellcolor{Gray0} 85.44$\pm$2.77 & \cellcolor{Gray0} 71.52$\pm$2.64 \\ 
 & \cellcolor{Gray1}  50 & \cellcolor{Gray1} 76.03$\pm$2.35 & \cellcolor{Gray1} 72.43$\pm$3.74 & \cellcolor{Gray1} 61.75$\pm$6.18 & \cellcolor{Gray1} 90.31$\pm$3.1 & \cellcolor{Gray1} 57.73$\pm$3.46 & \cellcolor{Gray1} 86.96$\pm$2.33 & \cellcolor{Gray1} 74.2$\pm$2.37 \\ 
 & \cellcolor{Gray2}  100 & \cellcolor{Gray2} 76.57$\pm$2.42 & \cellcolor{Gray2} 73.01$\pm$3.79 & \cellcolor{Gray2} 62.56$\pm$6.42 & \cellcolor{Gray2} 90.59$\pm$3.22 & \cellcolor{Gray2} 58.97$\pm$3.42 & \cellcolor{Gray2} 87.31$\pm$2.34 & \cellcolor{Gray2} 74.84$\pm$2.42 \\ 
 & \cellcolor{Gray3}  200 & \cellcolor{Gray3} 76.82$\pm$2.4 & \cellcolor{Gray3} 72.94$\pm$3.92 & \cellcolor{Gray3} 61.74$\pm$6.07 & \cellcolor{Gray3} 91.9$\pm$2.94 & \cellcolor{Gray3} 60.13$\pm$3.31 & \cellcolor{Gray3} 88.07$\pm$2.19 & \cellcolor{Gray3} 75.26$\pm$2.41 \\ \hline
\multirow{4}{*}{AdaC2}       
 & \cellcolor{Gray0}  25 & \cellcolor{Gray0} 78.06$\pm$3.79 & \cellcolor{Gray0} 75.4$\pm$5.53 & \cellcolor{Gray0} 67.53$\pm$10.34 & \cellcolor{Gray0} 88.6$\pm$4.74 & \cellcolor{Gray0} 58.43$\pm$3.35 & \cellcolor{Gray0} 88.16$\pm$2.38 & \cellcolor{Gray0} 76.03$\pm$3.39 \\ 
 & \cellcolor{Gray1}  50 & \cellcolor{Gray1} 76.65$\pm$3.29 & \cellcolor{Gray1} 72.9$\pm$5.46 & \cellcolor{Gray1} 61.04$\pm$9.1 & \cellcolor{Gray1} 92.26$\pm$4.1 & \cellcolor{Gray1} 59.75$\pm$3.75 & \cellcolor{Gray1} 88.48$\pm$2.57 & \cellcolor{Gray1} 75.18$\pm$3.22 \\ 
 & \cellcolor{Gray2}  100 & \cellcolor{Gray2} 77.69$\pm$2.82 & \cellcolor{Gray2} 74.05$\pm$4.91 & \cellcolor{Gray2} 61.79$\pm$7.86 & \cellcolor{Gray2} 93.59$\pm$3.15 & \cellcolor{Gray2} 61.95$\pm$3.37 & \cellcolor{Gray2} 89.18$\pm$2.1 & \cellcolor{Gray2} 76.37$\pm$2.82 \\ 
 & \cellcolor{Gray3}  200 & \cellcolor{Gray3} 78.25$\pm$2.46 & \cellcolor{Gray3} 74.22$\pm$4.21 & \cellcolor{Gray3} 62.03$\pm$6.75 & \cellcolor{Gray3} 94.48$\pm$2.3 & \cellcolor{Gray3} 63.04$\pm$3.15 & \cellcolor{Gray3} 89.92$\pm$1.88 & \cellcolor{Gray3} 76.99$\pm$2.4 \\ \hline
\multirow{4}{*}{AdaC3}    
 & \cellcolor{Gray0} 25 & \cellcolor{Gray0} 75.71$\pm$3.84 & \cellcolor{Gray0} 70.31$\pm$6.81 & \cellcolor{Gray0} 68.21$\pm$10.45 & \cellcolor{Gray0} 83.21$\pm$6.31 & \cellcolor{Gray0} 52.32$\pm$3.68 & \cellcolor{Gray0} 86.51$\pm$2.49 & \cellcolor{Gray0} 72.71$\pm$3.48 \\ 
 & \cellcolor{Gray1} 50 & \cellcolor{Gray1} 73.3$\pm$3.46 & \cellcolor{Gray1} 65.44$\pm$7.09 & \cellcolor{Gray1} 63.87$\pm$9.58 & \cellcolor{Gray1} 82.74$\pm$7.2 & \cellcolor{Gray1} 52.44$\pm$3.72 & \cellcolor{Gray1} 86.88$\pm$2.67 & \cellcolor{Gray1} 70.78$\pm$3.5 \\ 
 & \cellcolor{Gray2}  100 & \cellcolor{Gray2} 74.06$\pm$3.35 & \cellcolor{Gray2} 66.77$\pm$7.15 & \cellcolor{Gray2} 61.88$\pm$8.62 & \cellcolor{Gray2} 86.23$\pm$6.14 & \cellcolor{Gray2} 55.79$\pm$3.74 & \cellcolor{Gray2} 87.53$\pm$2.56 & \cellcolor{Gray2} 72.04$\pm$3.5 \\ 
 & \cellcolor{Gray3}  200 & \cellcolor{Gray3} 75.2$\pm$3.09 & \cellcolor{Gray3} 68.26$\pm$6.37 & \cellcolor{Gray3} 61.59$\pm$8.18 & \cellcolor{Gray3} 88.82$\pm$4.89 & \cellcolor{Gray3} 58.79$\pm$3.61 & \cellcolor{Gray3} 88.35$\pm$2.23 & \cellcolor{Gray3} 73.5$\pm$3.19 \\ \hline
\multirow{4}{*}{RareBoost}  
 & \cellcolor{Gray0} 25 & \cellcolor{Gray0} 75.83$\pm$1.99 & \cellcolor{Gray0} 68.44$\pm$3.57 & \cellcolor{Gray0} 54.94$\pm$4.07 & \cellcolor{Gray0} \circletext{96.72$\pm$0.48} & \cellcolor{Gray0} 60.03$\pm$3.39 & \cellcolor{Gray0} 81.96$\pm$2.79 & \cellcolor{Gray0} 72.99$\pm$2.31 \\ 
 & \cellcolor{Gray1}  50 & \cellcolor{Gray1} 77.51$\pm$1.83 & \cellcolor{Gray1} 70.96$\pm$3.14 & \cellcolor{Gray1} 58.17$\pm$3.69 & \cellcolor{Gray1} \circletext{96.84$\pm$0.39} & \cellcolor{Gray1} 62.82$\pm$3.09 & \cellcolor{Gray1} 78.64$\pm$2.78 & \cellcolor{Gray1} 74.16$\pm$2.09 \\ 
 & \cellcolor{Gray2} 100 & \cellcolor{Gray2} 78.55$\pm$1.78 & \cellcolor{Gray2} 72.47$\pm$3.08 & \cellcolor{Gray2} 60.19$\pm$3.6 & \cellcolor{Gray2} \circletext{96.91$\pm$0.39} & \cellcolor{Gray2} 64.44$\pm$3.01 & \cellcolor{Gray2} 75.02$\pm$2.59 & \cellcolor{Gray2} 74.6$\pm$2.05 \\ 
 & \cellcolor{Gray3}  200 & \cellcolor{Gray3} 79.04$\pm$1.69 & \cellcolor{Gray3} 73.29$\pm$2.94 & \cellcolor{Gray3} 61.16$\pm$3.4 & \cellcolor{Gray3} \circletext{96.93$\pm$0.4} & \cellcolor{Gray3} 65.18$\pm$2.96 & \cellcolor{Gray3} 71.39$\pm$2.46 & \cellcolor{Gray3} 74.5$\pm$1.96 \\ \hline
\end{tabular}\end{adjustbox}
\end{table}

\begin{table}[tp!]\centering
\caption{Results for various evaluation metrics for the comparison of AdaCC1/2 versus AdaN-CC1/2. Best and second best methods per different values of $T$ are in bold and circled, respectively. Colors indicate specific values of T.}
\label{tbl:counter_parts_comparison} 
\begin{adjustbox}{width=1\textwidth}
\begin{tabular}{ccccccccc}
\hline
Method                       & T & Bal. Acc & Gmean & TPR (Recall) & TNR & F1Score & AUC & OPM \\ \hline
\multirow{4}{*}{AdaCC1}      
 & \cellcolor{Gray0} 25 & \cellcolor{Gray0} \textbf{83.16$\pm$3.29} & \cellcolor{Gray0} \circletext{82.11$\pm$6.23} & \cellcolor{Gray0} \circletext{79.68$\pm$8.42} & \cellcolor{Gray0} 86.65$\pm$3.65 & \cellcolor{Gray0} \circletext{55.8$\pm$4.86} & \cellcolor{Gray0} \textbf{90.28$\pm$1.83} & \cellcolor{Gray0} \textbf{79.61$\pm$3.57} \\ 
 & \cellcolor{Gray1}  50 & \cellcolor{Gray1} \textbf{84.25$\pm$2.64} & \cellcolor{Gray1} \circletext{83.53$\pm$4.56} & \cellcolor{Gray1} \textbf{81.92$\pm$4.71} & \cellcolor{Gray1} 86.58$\pm$4.83 & \cellcolor{Gray1} \circletext{58.16$\pm$3.51} & \cellcolor{Gray1} \textbf{90.98$\pm$1.6} & \cellcolor{Gray1} \textbf{80.9$\pm$2.74} \\ 
 & \cellcolor{Gray2}  100 & \cellcolor{Gray2} \textbf{85.01$\pm$2.07} & \cellcolor{Gray2} \textbf{84.6$\pm$2.8} & \cellcolor{Gray2} \textbf{81.67$\pm$3.75} & \cellcolor{Gray2} 88.35$\pm$2.2 & \cellcolor{Gray2} \textbf{60.48$\pm$2.73} & \cellcolor{Gray2} \textbf{91.49$\pm$1.48} & \cellcolor{Gray2} \textbf{81.93$\pm$2.02} \\ 
 & \cellcolor{Gray3} 200 & \cellcolor{Gray3} \textbf{85.21$\pm$1.85} & \cellcolor{Gray3} \textbf{84.64$\pm$2.41} & \cellcolor{Gray3} \textbf{81.19$\pm$3.52} & \cellcolor{Gray3} 89.23$\pm$1.55 & \cellcolor{Gray3} \circletext{61.91$\pm$2.8} & \cellcolor{Gray3} \textbf{91.78$\pm$1.38} & \cellcolor{Gray3} \textbf{82.32$\pm$1.83} \\ \hline
\multirow{4}{*}{AdaCC2}      
 & \cellcolor{Gray0}  25 & \cellcolor{Gray0} \circletext{82.97$\pm$2.06} & \cellcolor{Gray0} \textbf{82.47$\pm$2.38} & \cellcolor{Gray0} \textbf{80.69$\pm$5.79} & \cellcolor{Gray0} 85.25$\pm$3.53 & \cellcolor{Gray0} \textbf{56.24$\pm$3.04} & \cellcolor{Gray0} \circletext{89.76$\pm$1.91} & \cellcolor{Gray0} \circletext{79.55$\pm$1.93} \\ 
 & \cellcolor{Gray1}  50 & \cellcolor{Gray1} \circletext{84.17$\pm$1.98} & \cellcolor{Gray1} \textbf{83.66$\pm$2.33} & \cellcolor{Gray1} \circletext{80.69$\pm$5.25} & \cellcolor{Gray1} \circletext{87.65$\pm$2.85} & \cellcolor{Gray1} \textbf{58.9$\pm$3.0} & \cellcolor{Gray1} \circletext{89.72$\pm$2.2} & \cellcolor{Gray1} \circletext{80.79$\pm$1.96} \\ 
 & \cellcolor{Gray2} 100 & \cellcolor{Gray2} \circletext{84.46$\pm$1.95} & \cellcolor{Gray2} \circletext{83.71$\pm$2.42} & \cellcolor{Gray2} \circletext{80.54$\pm$5.04} & \cellcolor{Gray2} \circletext{88.38$\pm$2.82} & \cellcolor{Gray2} \circletext{60.08$\pm$3.16} & \cellcolor{Gray2} \circletext{89.16$\pm$2.58} & \cellcolor{Gray2} \circletext{81.05$\pm$1.99} \\ 
 & \cellcolor{Gray3}  200 & \cellcolor{Gray3} \circletext{84.41$\pm$1.85} & \cellcolor{Gray3} \circletext{83.31$\pm$2.37} & \cellcolor{Gray3} \circletext{79.22$\pm$4.76} & \cellcolor{Gray3} \circletext{89.6$\pm$2.36} & \cellcolor{Gray3} \textbf{62.4$\pm$3.01} & \cellcolor{Gray3} \circletext{87.67$\pm$3.56} & \cellcolor{Gray3} \circletext{81.1$\pm$2.0} \\ \hline
\multirow{4}{*}{AdaN-CC1}    
 & \cellcolor{Gray0} 25 & \cellcolor{Gray0} 70.79$\pm$4.59 & \cellcolor{Gray0} 52.52$\pm$10.63 & \cellcolor{Gray0} 48.74$\pm$11.07 & \cellcolor{Gray0} \circletext{92.83$\pm$3.14} & \cellcolor{Gray0} 41.26$\pm$8.5 & \cellcolor{Gray0} 82.06$\pm$5.19 & \cellcolor{Gray0} 64.7$\pm$5.92 \\ 
 & \cellcolor{Gray1} 50 & \cellcolor{Gray1} 72.57$\pm$4.47 & \cellcolor{Gray1} 58.7$\pm$9.7 & \cellcolor{Gray1} 59.45$\pm$14.94 & \cellcolor{Gray1} 85.68$\pm$11.41 & \cellcolor{Gray1} 43.97$\pm$8.48 & \cellcolor{Gray1} 79.56$\pm$6.36 & \cellcolor{Gray1} 66.66$\pm$5.58 \\ 
 & \cellcolor{Gray2} 100 & \cellcolor{Gray2} 72.6$\pm$4.51 & \cellcolor{Gray2} 58.73$\pm$9.73 & \cellcolor{Gray2} 58.75$\pm$13.64 & \cellcolor{Gray2} 86.45$\pm$9.63 & \cellcolor{Gray2} 43.93$\pm$8.44 & \cellcolor{Gray2} 77.23$\pm$7.16 & \cellcolor{Gray2} 66.28$\pm$5.49 \\ 
 & \cellcolor{Gray3}  200 & \cellcolor{Gray3} 72.6$\pm$4.51 & \cellcolor{Gray3} 58.73$\pm$9.73 & \cellcolor{Gray3} 58.46$\pm$13.5 & \cellcolor{Gray3} 86.75$\pm$8.87 & \cellcolor{Gray3} 43.89$\pm$8.46 & \cellcolor{Gray3} 76.52$\pm$7.27 & \cellcolor{Gray3} 66.16$\pm$5.43 \\ \hline
\multirow{4}{*}{AdaN-CC2}   
 & \cellcolor{Gray0}  25 & \cellcolor{Gray0} 75.23$\pm$3.37 & \cellcolor{Gray0} 66.04$\pm$7.1 & \cellcolor{Gray0} 55.21$\pm$8.1 & \cellcolor{Gray0} \textbf{95.25$\pm$2.1} & \cellcolor{Gray0} 53.21$\pm$5.9 & \cellcolor{Gray0} 84.88$\pm$4.66 & \cellcolor{Gray0} 71.64$\pm$4.11 \\ 
 & \cellcolor{Gray1}  50 & \cellcolor{Gray1} 76.34$\pm$3.58 & \cellcolor{Gray1} 67.14$\pm$7.83 & \cellcolor{Gray1} 61.8$\pm$8.91 & \cellcolor{Gray1} \textbf{90.88$\pm$3.29} & \cellcolor{Gray1} 56.06$\pm$6.06 & \cellcolor{Gray1} 79.81$\pm$7.98 & \cellcolor{Gray1} 72.0$\pm$4.45 \\ 
 & \cellcolor{Gray2}  100 & \cellcolor{Gray2} 76.51$\pm$3.72 & \cellcolor{Gray2} 67.41$\pm$8.07 & \cellcolor{Gray2} 62.15$\pm$9.19 & \cellcolor{Gray2} \textbf{90.86$\pm$3.28} & \cellcolor{Gray2} 56.34$\pm$6.33 & \cellcolor{Gray2} 75.28$\pm$9.63 & \cellcolor{Gray2} 71.43$\pm$4.54 \\ 
 & \cellcolor{Gray3}   200 & \cellcolor{Gray3} 76.5$\pm$3.72 & \cellcolor{Gray3} 67.41$\pm$8.07 & \cellcolor{Gray3} 62.15$\pm$9.19 & \cellcolor{Gray3} \textbf{90.86$\pm$3.28} & \cellcolor{Gray3} 56.34$\pm$6.33 & \cellcolor{Gray3} 74.9$\pm$9.67 & \cellcolor{Gray3} 71.36$\pm$4.51 \\ \hline
\end{tabular}\end{adjustbox}
\end{table}

\begin{table}[htp!]
\centering
\caption{Comparative Balanced Accuracy ranks across the entire set of methods and datasets (smaller values are better) for $T=200$. Best methods per dataset are in bold. Last row (winner) indicates on how many datasets a method is ranked first (best balanced accuracy score, higher values are better). Note that in some datasets, the methods have equal scores (tie); therefore, the ranks are in float format.}
\label{tbl:ranks}
\begin{adjustbox}{width=1\textwidth}
\begin{tabular}{lcccccccccccccc}
\hline
 & AdaBoost & AdaCC1 & AdaCC2 & AdaMEC & AdaMEC-Cal. & CGAda & CGAda-Cal. & AdaCost & CSB1 & CSB2 & AdaC1 & AdaC2 & AdaC3 & RareBoost \\ \hline
abalone & 14.0 & 3.0 & 4.0 & 2.0 & 7.0 & 9.0 & 8.0 & \textbf{1.0} & 11.0 & 13.0 & 5.0 & 6.0 & 10.0 & 12.0 \\
adult & 9.5 & \textbf{1.0} & 2.0 & 9.5 & 5.0 & 4.0 & 3.0 & 7.0 & 14.0 & 9.5 & 12.0 & 9.5 & 13.0 & 6.0 \\
bank & 12.5 & \textbf{1.0} & 3.0 & 2.0 & 6.0 & 4.0 & 5.0 & 7.0 & 14.0 & 12.5 & 9.0 & 11.0 & 8.0 & 10.0 \\
car eval. & 9.5 & \textbf{1.0} & 3.0 & 9.5 & 5.0 & 6.0 & 4.0 & 14.0 & 12.0 & 9.5 & 2.0 & 9.5 & 13.0 & 7.0 \\
coil 2000 & 13.0 & \textbf{1.0} & 3.0 & 2.0 & 8.0 & 5.0 & 7.0 & 9.0 & 10.0 & 14.0 & 4.0 & 6.0 & 12.0 & 11.0 \\
credit & 13.5 & 2.5 & 8.0 & 5.0 & \textbf{1.0} & 4.0 & 2.5 & 7.0 & 11.0 & 13.5 & 6.0 & 10.0 & 9.0 & 12.0 \\
eeg eye & 5.0 & 2.0 & 11.0 & 12.0 & 3.0 & 8.0 & 7.0 & 14.0 & 13.0 & 5.0 & 10.0 & 5.0 & 9.0 & \textbf{1.0} \\
electricity & 7.5 & 2.5 & 10.0 & 7.5 & 4.0 & 2.5 & 5.0 & 13.0 & 14.0 & 7.5 & 12.0 & 7.5 & 11.0 & \textbf{1.0} \\
isolet & 9.5 & \textbf{1.0} & 2.0 & 9.5 & 4.0 & 7.0 & 5.0 & 14.0 & 3.0 & 9.5 & 12.0 & 9.5 & 13.0 & 6.0 \\
letter img. & 8.5 & 2.0 & \textbf{1.0} & 8.5 & 4.0 & 5.0 & 6.0 & 13.0 & 14.0 & 8.5 & 11.5 & 8.5 & 11.5 & 3.0 \\
mammography & 9.5 & 2.0 & \textbf{1.0} & 9.5 & 3.0 & 5.0 & 6.0 & 4.0 & 14.0 & 9.5 & 12.0 & 9.5 & 13.0 & 7.0 \\
musk2 & 8.5 & 3.0 & 2.0 & 8.5 & 4.0 & 6.0 & 5.0 & 11.0 & 14.0 & 8.5 & 12.5 & 8.5 & 12.5 & \textbf{1.0} \\
optical digits & 8.5 & \textbf{1.0} & 2.0 & 8.5 & 5.0 & 6.0 & 4.0 & 14.0 & 13.0 & 8.5 & 11.5 & 8.5 & 11.5 & 3.0 \\
ozone level & 11.5 & 2.0 & 3.0 & 11.5 & 4.0 & 6.0 & 5.0 & \textbf{1.0} & 7.0 & 11.5 & 14.0 & 11.5 & 8.0 & 9.0 \\
pen digits & 8.5 & 2.0 & \textbf{1.0} & 8.5 & 6.0 & 4.0 & 5.0 & 14.0 & 13.0 & 8.5 & 11.5 & 8.5 & 11.5 & 3.0 \\
phoneme & 9.5 & 5.0 & 4.0 & 7.0 & 2.0 & 1.0 & 3.0 & 12.0 & 14.0 & 9.5 & 11.0 & 6.0 & 13.0 & 8.0 \\
protein hom. & 8.5 & 2.0 & \textbf{1.0} & 8.5 & 3.0 & 4.0 & 6.0 & 14.0 & 13.0 & 8.5 & 11.5 & 8.5 & 11.5 & 5.0 \\
satimage & 12.0 & 2.0 & \textbf{1.0} & 10.0 & 4.0 & 5.0 & 3.0 & 9.0 & 14.0 & 12.0 & 8.0 & 12.0 & 7.0 & 6.0 \\
scene & 13.5 & \textbf{1.0} & 7.0 & 9.0 & 5.0 & 4.0 & 6.0 & 3.0 & 8.0 & 13.5 & 2.0 & 12.0 & 11.0 & 10.0 \\
sick euthyroid & 9.5 & 2.0 & \textbf{1.0} & 9.5 & 7.0 & 4.0 & 6.0 & 12.0 & 14.0 & 9.5 & 3.0 & 9.5 & 5.0 & 13.0 \\
skin & 8.5 & 5.0 & 4.0 & 8.5 & 2.5 & 1.0 & 2.5 & 13.0 & 14.0 & 8.5 & 11.5 & 8.5 & 11.5 & 6.0 \\
spam & 8.5 & 4.5 & 2.0 & 8.5 & 4.5 & 6.0 & 3.0 & 14.0 & 13.0 & 8.5 & 11.5 & 8.5 & 11.5 & \textbf{1.0} \\
thyroid sick & 10.0 & \textbf{1.0} & 2.0 & 8.0 & 4.0 & 6.0 & 5.0 & 3.0 & 14.0 & 10.0 & 12.0 & 10.0 & 13.0 & 7.0 \\
us crime & 9.5 & \textbf{1.0} & 4.0 & 9.5 & 6.0 & 7.0 & 5.0 & 3.0 & 2.0 & 9.5 & 13.0 & 9.5 & 12.0 & 14.0 \\
webpage & 11.5 & 2.0 & \textbf{1.0} & 3.0 & 5.0 & 7.0 & 6.0 & 8.0 & 9.0 & 11.5 & 14.0 & 10.0 & 13.0 & 4.0 \\
wilt & 9.5 & 2.0 & \textbf{1.0} & 9.5 & 3.0 & 5.0 & 4.0 & 14.0 & 6.0 & 9.5 & 12.5 & 9.5 & 12.5 & 7.0 \\
wine quality & 13.5 & \textbf{1.0} & 2.0 & 3.0 & 5.0 & 7.0 & 6.0 & 4.0 & 12.0 & 13.5 & 8.0 & 9.0 & 10.0 & 11.0 \\ \hline
avg. & 10.11 & \textbf{2.06} & 3.19 & 7.70 & 4.44 & 5.13 & 4.93 & 9.33 & 11.48 & 10.11 & 9.74 & 8.96 & 11.00 & 6.81 \\ \hline
winner & 0 & \textbf{10} & 8 & 0 & 1 & 2 & 0 & 2 & 0 & 0 & 0 & 0 & 0 & 4 \\ \hline
\end{tabular}\end{adjustbox}
\end{table}

\begin{table}[htp!]\centering
\caption{Friedman test: p-values for all competitors. Non-significant values ($p > 0.05$) are in bold. Colors indicate specific values of $T=[25,50,100,200]$.}
\label{tbl:sig_test}
\begin{adjustbox}{width=1\textwidth}
\begin{tabular}{labde|abde}
\hline
 & \multicolumn{4}{c}{AdaCC1} & \multicolumn{4}{c|}{AdaCC2} \\ 
$T =$ & 25 & 50 & 100 & 200 & 25 &  50 &  100 & 200 \\\hline 
AdaBoost & 0.0e+00 & 3.6e-14 & 6.5e-13 & 4.0e-11 & 6.6e-15 & 4.9e-14 & 8.5e-12 & 1.1e-08 \\
AdaCC1 & - & - & - & - & \textbf{7.7e-01} & \textbf{9.6e-01} & \textbf{7.3e-01} & \textbf{3.9e-01} \\
AdaCC2 & \textbf{7.7e-01} & \textbf{9.6e-01} & \textbf{7.3e-01} & \textbf{3.9e-01} & - & - & - & - \\
AdaMEC & 2.7e-03 & 3.7e-04 & 8.7e-06 & 7.6e-06 & 7.5e-03 & 4.5e-04 & 4.6e-05 & 3.7e-04 \\
AdaMEC-Cal. & 1.2e-03 & 6.1e-03 & \textbf{8.7e-02} & \textbf{7.6e-02} & 2.9e-02 & 6.8e-03 & 1.8e-02 & \textbf{4.3e-01} \\
CGAda & 2.8e-03 & 3.4e-02 & 2.2e-02 & 3.1e-02 & 7.7e-03 & 3.8e-02 & 4.0e-02 & 4.7e-02\\
CGAda-Cal. & 4.7e-03 & 3.4e-02 & 2.2e-02 & \textbf{8.0e-02} & 1.2e-02 & 3.8e-02 & 4.0e-02 & \textbf{3.0e-01} \\
AdaCost & 2.6e-09 & 9.5e-09 & 1.0e-09 & 2.1e-09 & 1.6e-08 & 1.2e-08 & 9.5e-09 & 3.6e-07 \\
CSB1 & 7.8e-07 & 1.1e-07 & 2.7e-12 & 3.3e-15 & 3.6e-06 & 1.4e-07 & 3.3e-11 & 3.0e-12 \\
CSB2 & 5.7e-06 & 1.6e-13 & 5.6e-13 & 3.6e-11 & 2.3e-05 & 2.3e-13 & 7.4e-12 & 1.0e-08 \\
AdaC1 & 2.9e-09 & 9.5e-09 & 1.0e-09 & 3.6e-10 & 1.7e-08 & 1.2e-08 & 9.5e-09 & 7.7e-08 \\
AdaC2 & 5.7e-06 & 8.0e-09 & 8.3e-09 & 2.9e-08 & 2.3e-05 & 1.0e-08 & 6.7e-08 & 3.4e-06 \\
AdaC3 & 1.9e-09 & 5.0e-12 & 3.4e-13 & 1.1e-13 & 1.2e-08 & 6.8e-12 & 4.6e-12 & 6.4e-11 \\
RareBoost & 6.976e-08 & 6.581e-06 & 1.502e-04 & 1.8e-04 & 3.6e-07 & 8.1e-06 & 6.4e-04 & 5.0e-03 \\ \hline
\end{tabular}
\end{adjustbox}
\end{table}

%% file: exp_internal_behavior.tex
We begin the internal analysis by comparing our methods, AdaCC1 and AdaCC2, with their corresponding non-cumulative version, 
namely AdaN-CC1 and AdaN-CC2, which are introduced in Section~\ref{sec:baselines}. Then, we continue our analysis 
in which we compare our methods with competitors w.r.t. in-training instance re-weighting, $\alpha$ estimation, feature importance, confidence scores and decision boundaries (similar to the toy example in Figure~\ref{fig:toy_example}).

\begin{figure*}   \centering
  \begin{subfigure}[t]{.477\textwidth}
 \includegraphics[width=1.0\columnwidth]{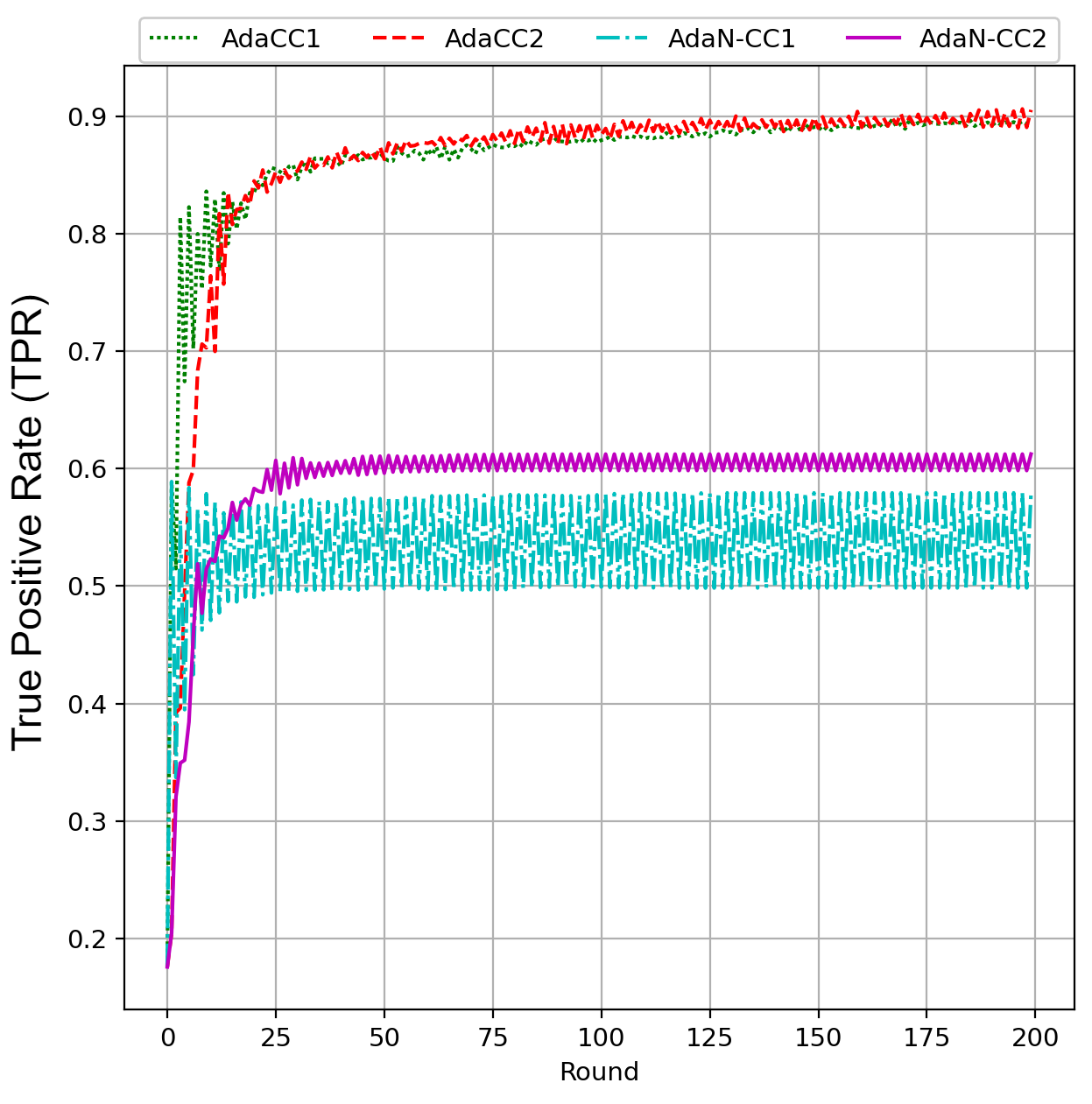}
 \caption{In-training TPR per boosting round}
 \label{fig:a_v_na_tpr}
 \end{subfigure}
  \begin{subfigure}[t]{.49\textwidth}
  \centering
 \includegraphics[width=1.0\columnwidth]{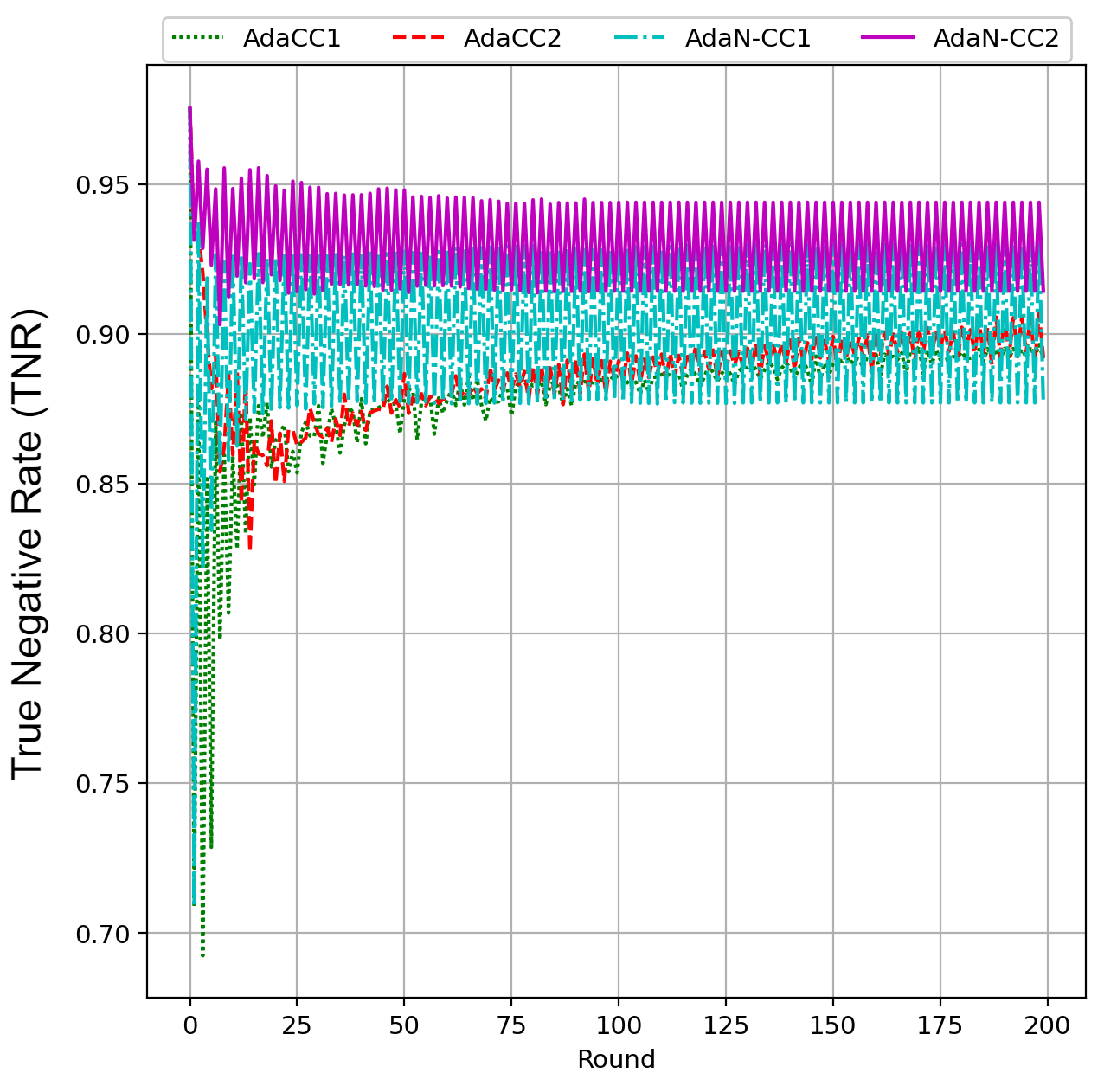}
 \caption{In-training TNR per boosting round} 
 \label{fig:a_v_na_tnr}
 \end{subfigure}
 \caption{Cumulative vs non-cumulative misclassification cost estimation (left:TPR, right:TNR).}
 \label{fig:amort_vs_non_amort}
\end{figure*}

\noindent\textbf{Cumulative vs Non-Cumulative Costs:} In Figure~\ref{fig:amort_vs_non_amort} we compare AdaCC1/2 and AdaN-CC1/2 on the TPR and TNR values per boosting round (averaged over the datasets, $T=200$). Figure~\ref{fig:a_v_na_tpr} shows the in-training TPR scores over the boosting rounds. It is clear that the cumulative versions, i.e., AdaCC1 and AdaCC2, are by far better and more stable than the non-cumulative ones, AdaN-CC1 and AdaN-CC2. Figure~\ref{fig:a_v_na_tnr} shows the in-training TNR scores over the boosting rounds. The non-cumulative versions are better than the cumulative ones. However, they exhibit high fluctuation as they rely on point-in-time estimates of misclassification costs (i.e., based on individual weak learners) comparing to the cumulative methods which rely on cumulative estimates (i.e., based on the partial ensemble). 
These experiments demonstrate the importance of the cumulative misclassification cost estimation for the stability of the model. Also, in terms of predictive performance, we have seen (c.f., Table~\ref{tbl:results}) that the non-cumulative methods, AdaN-CC1 and AdaN-CC2, are producing significantly worse results in contrast to AdaCC1 and AdaCC2.

\begin{figure*}[ht!]
 \centering
 \begin{subfigure}[t]{.48\textwidth}
 \centering
 \includegraphics[width=1.0\columnwidth]{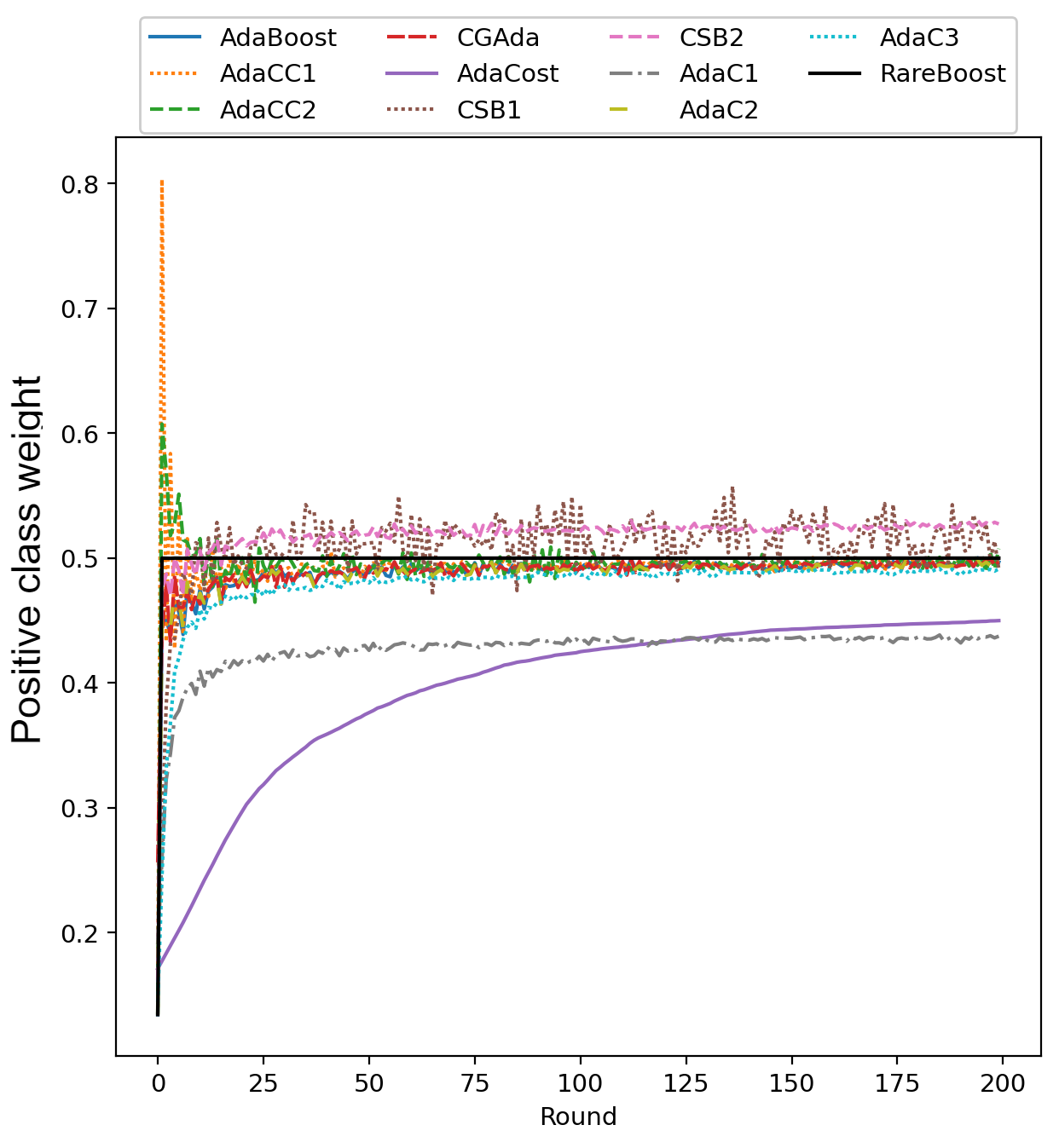}
 \caption{Positive (minority) class weight}
 \label{fig:in_training_comparisons_positive}
 \end{subfigure} 
 \centering
 \begin{subfigure}[t]{.48\textwidth}
 \centering
\includegraphics[width=1.0\columnwidth]{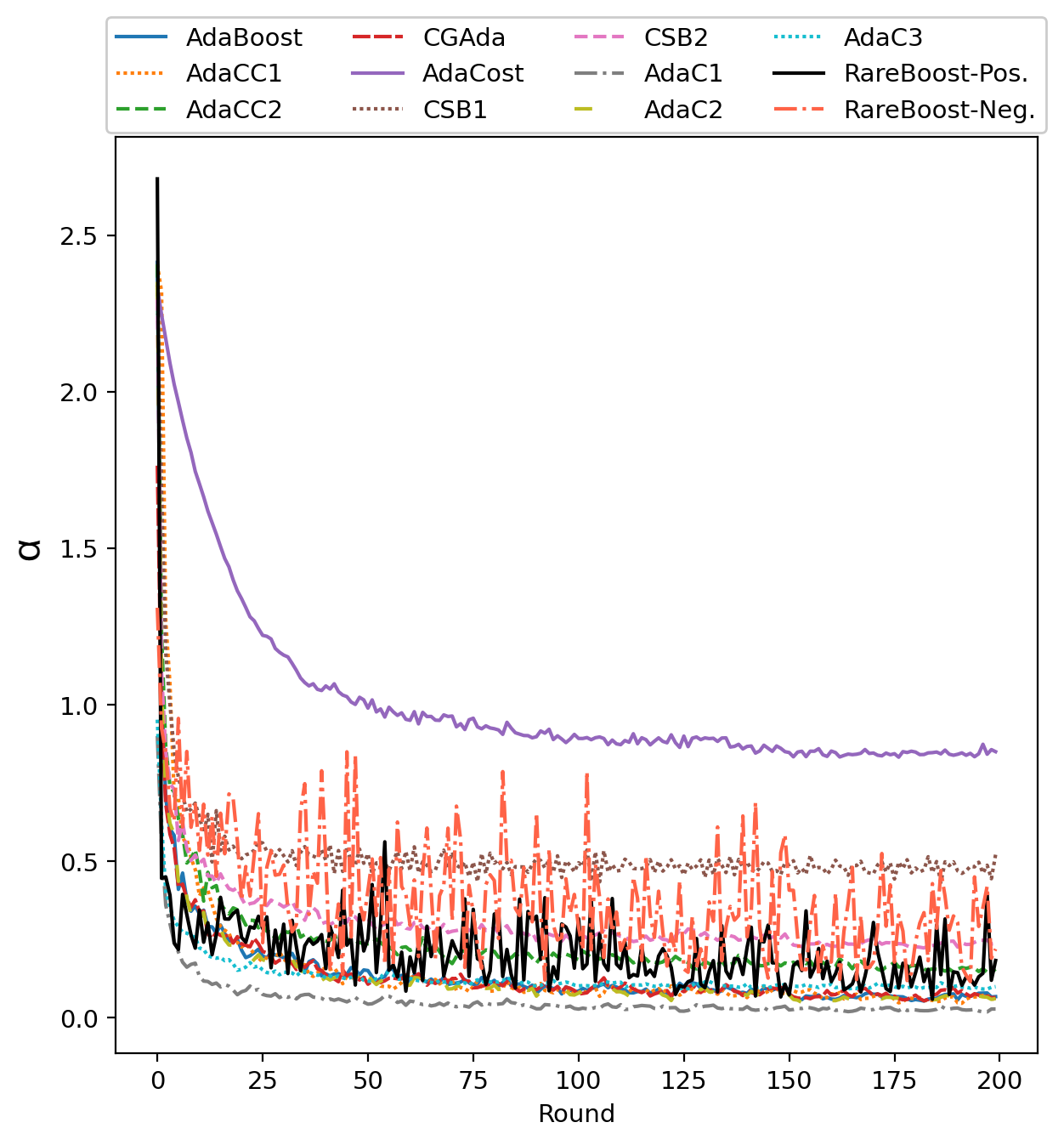}
 \caption{Alpha value ($\alpha$)} 
 \label{fig:in_training_comparisons_apha}
 \end{subfigure}
 \begin{subfigure}[t]{.48\textwidth}
 \centering
 \includegraphics[width=1.0\columnwidth]{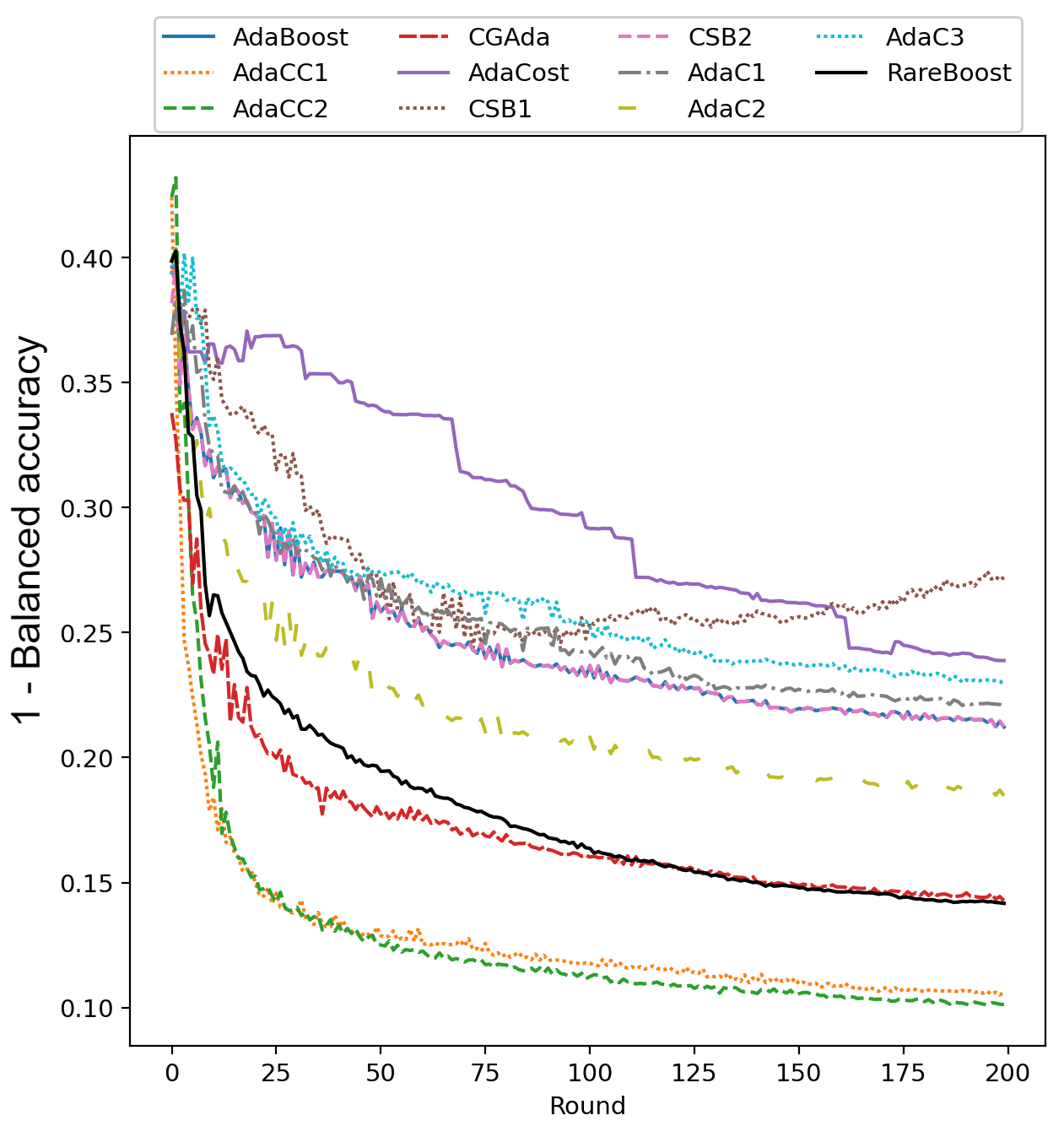}
 \caption{In-training balanced error} 
 \label{fig:in_training_comparisons_error}
 \end{subfigure}
 \caption{In-training behavior over the boosting rounds (for $T=200$).} 
 \label{fig:in_training_comparisons}
\end{figure*}

\begin{figure}[tp!]
 \centering
 \includegraphics[width=.7\textwidth]{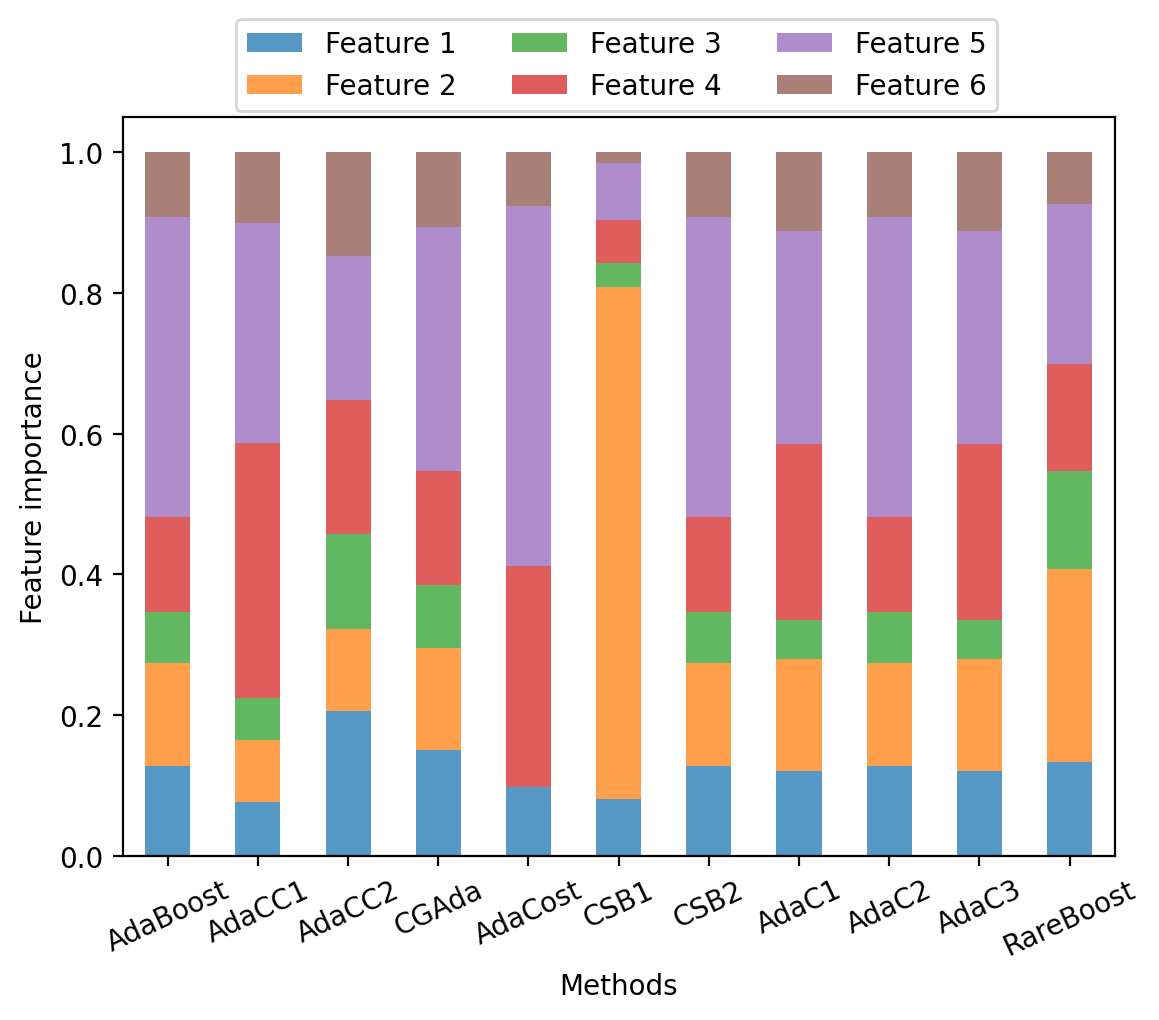}
 \caption{Feature importance of mammography dataset (the higher, the more important the feature).}
 \label{fig:feature_importance}
\end{figure}

\begin{figure*}[tp!]
 \centering
 \begin{subfigure}[t]{1\textwidth}
 \centering
 \includegraphics[width=1\textwidth]{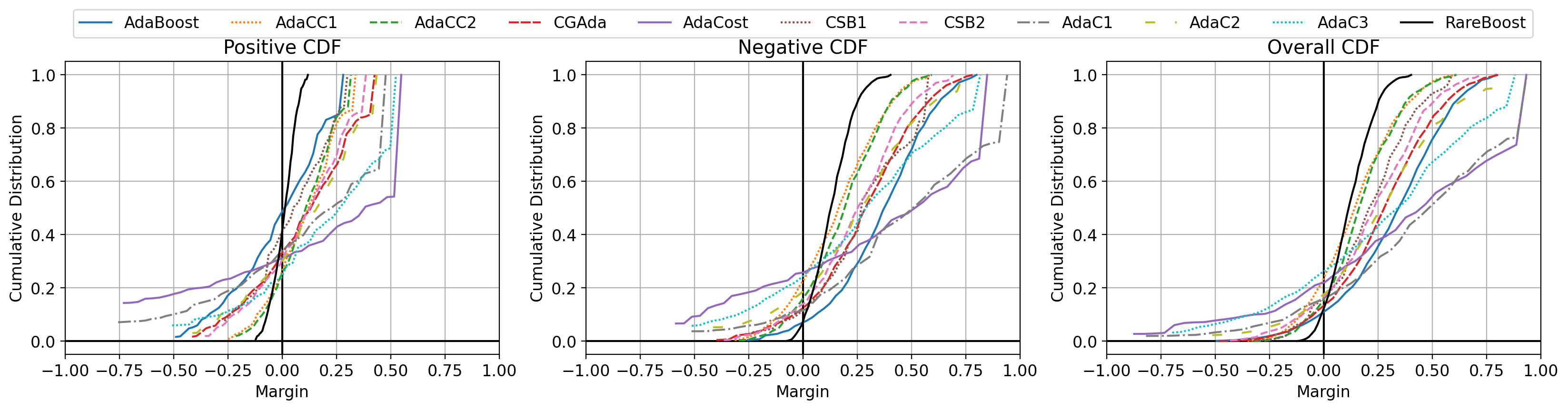}
 \caption{T = 25}
 \label{fig:cdf_t_25}
 \end{subfigure}
 \centering
 \begin{subfigure}[t]{1\textwidth}
 \centering
 \includegraphics[width=1\textwidth]{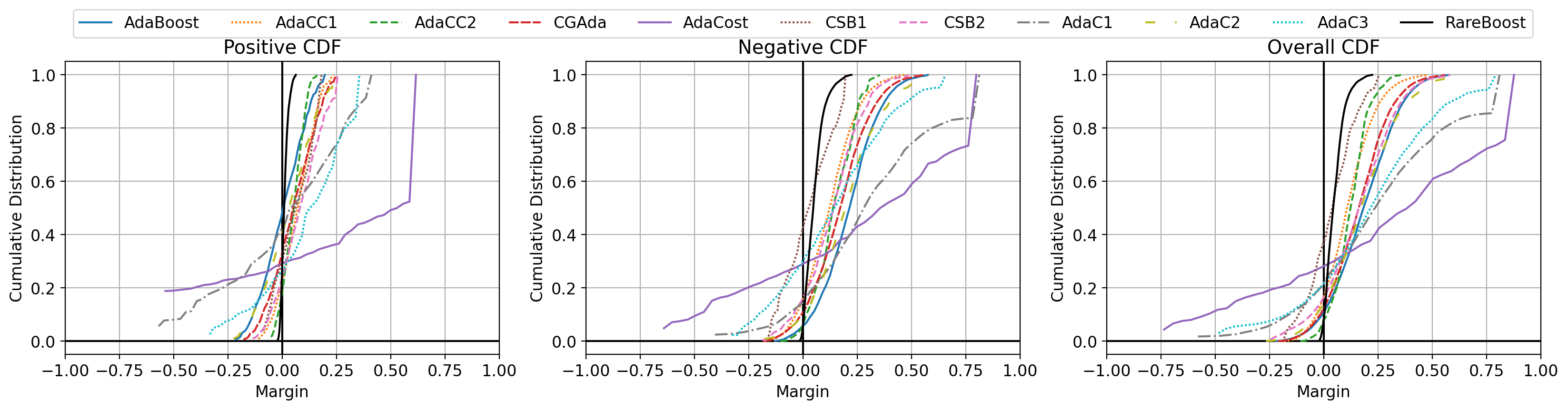}
 \caption{T = 200}
 \label{fig:cdf_t_200}
 \end{subfigure}
 \caption{Effect of boosting rounds on the confidence scores (left:positive class, middle:negative class, left:overall).}
 \label{fig:cdf_plots}
\end{figure*}

\begin{figure}[htp!]
 \centering
 \includegraphics[width=1\textwidth]{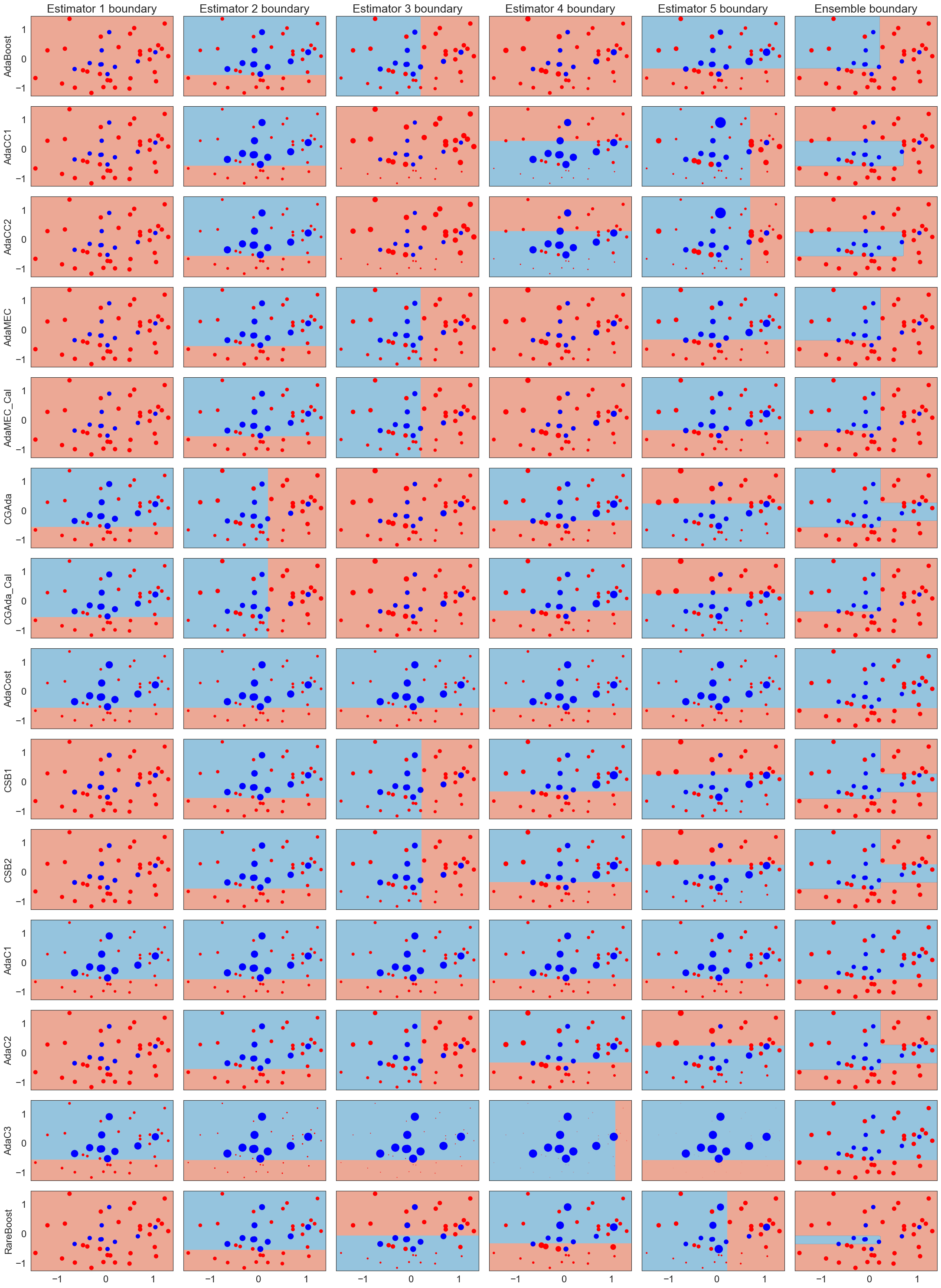}
 \caption{Decision boundaries of methods on the same imbalanced toy dataset of 10 blue and 30 red instances. Dot size is proportional to the weight allocated by each weak learner to the particular instance, making clear how each method assigns weights to minority class instances compared to the ones of the majority class.} 
 \label{fig:boundaries}
\end{figure}

\noindent\textbf{Model Performance Analysis:} The experiments thus far demonstrate the superior behavior of AdaCC1 and AdaCC2, compared to state-of-the-art cost-sensitive boosting approaches. Hereafter, we explain this behavior through additional experiments on the internal behavior of the models, assessed by: 
i) positive (minority) class weight assignments over the boosting rounds (Figure~\ref{fig:in_training_comparisons_positive}), 
ii) alpha values over the boosting rounds (Figure~\ref{fig:in_training_comparisons_apha}), iii) in-training balanced error over the boosting rounds (Figure~\ref{fig:in_training_comparisons_error}), iv) feature importance (Figure~\ref{fig:feature_importance}) of a given dataset (mammography), iv) confidence scores (Figure~\ref{fig:cdf_plots}), and v) decision boundaries (Figure~\ref{fig:boundaries}). Moreover, AdaMEC, AdaMEC-Cal., CGAda-Cal. are omitted from these experiments (except the decision boundary analysis). The reason is that AdaMEC is built on top of a trained AdaBoost model, by shifting its decision boundary towards the target class. AdaMEC-Cal. and CGAda-Cal. are calibrated versions of AdaMEC and CGAda.

\noindent\textbf{In-training Analysis:} 
For in-training analysis, we set $T=200$ and show the behavior of each method per boosting round. 
The weights of the minority class over the boosting rounds are shown in Figure~\ref{fig:in_training_comparisons_positive}; as we can see, AdaCC1 and AdaCC2 behave differently from the competitors by starting with very high weights during the first boosting rounds, which converge afterwards to $0.5$. The other methods increase the positive weights gradually over the rounds. Our methods tackle the class imbalance problem during early boosting rounds by assigning cumulative misclassifications costs to the minority class and then proceed to reduce these costs (dynamically) as soon as the TPR scores are close to TNR scores.

In terms of $\alpha$ values, which control how much the weak learners contribute to the final ensemble (Figure~\ref{fig:in_training_comparisons_apha}),
the methods depict a similar behavior with $\alpha$ decreasing over the boosting rounds. 
A notable exception is RareBoost which utilizes positive and negative $\alpha$ to estimate the weight distribution per round; thus, it is expected for its $\alpha$ values to fluctuate. Our methods do not differentiate from other competitors (excluding RareBoost); weak learners in the early boosting rounds (e.g., $T <10$) are more influential to the final outcomes (higher $\alpha$ values).  


In Figure~\ref{fig:in_training_comparisons_error} the in-training balanced error over the boosting rounds is shown. As we can see, our methods achieve the lowest error. 
Moreover, AdaCC1 and AdaCC2 reduce the balanced error faster than any other method, and converge after a sufficient number of boosting rounds. The abrupt reduction of the error is directly related to the rapid increase of the positive weights in the initial boosting rounds. 

\noindent\textbf{Feature Importance:} In Figure~\ref{fig:feature_importance} we illustrate the feature importance for each method on the mammography dataset. We have selected this dataset since it has low dimensionality (6 features) and high class imbalance ratio (1:42). Figure~\ref{fig:feature_importance} shows the importance of each feature which is employed by each method to make a decision (weights are normalized to be a distribution). Note that each weak learner is a decision stump which means that it selects only one feature for splitting the dataset. 
\rev{The feature importance is measured as follows: each ensemble consists of T weak learners and each weak learner is trained on a different data distribution. Since we have employed Decision Stumps (Decision trees of depth 1), each weak learner will use only one split; therefore, it will use only one feature. The weak learners of AdaCost, based on the data distributions which are provided (based on the model’s updating strategy), do not use some features based on the splitting criterion. In addition, some models (e.g., AdaCost) may terminate their boosting rounds earlier than others based on their stopping criterion which can lead to ignoring some features.~} Although the feature importance does not indicate which method is the best, it shows clearly that each method utilizes differently the features based on the weighting strategy e.g., AdaCC1 is relying more on features 4 and 5 and less on features 1 and 3 compared to AdaCC2.

\noindent\textbf{Confidence Analysis:} In Figure~\ref{fig:cdf_plots}, we compare the confidence scores of the different methods for two ensemble sizes, $T$=25 (Figure~\ref{fig:cdf_t_25}) and $T=200$ (Figure~\ref{fig:cdf_t_200}), and separate them into three categories: positive (left), negative (middle) and overall (right) confidence scores. Note that misclassified instances have confidence scores less than 0 on $x$-axis (values closer to 0, on $x$-axis, indicate lower confidence in the predictions, correct or wrong). Also, the area under the line in the range $[-1, 0]$ on the $x$-axis shows the proportion of misclassified instances.

At a first look at the overall confidence scores, we see that AdaCC1 and AdaCC2 are producing low misclassification rates while the area under the line in the $[-1,0]$ range of $x$-axis is low. However, other methods are achieving similar results. Therefore, we need to analyze the confidence scores of each class separately since the minority (positive) class is overshadowed by the majority (negative) class. 
As expected, AdaBoost has the highest misclassification confidence score in positive (minority) class since it learns effectively only the negative (majority) class. AdaCC1 and AdaCC2 methods have the lowest misclassification confidence scores for $T=25$, and reduce them even more as the number of weak learners increases, i.e., for $T=200$.  
For the negative class, our approaches are able to reduce the misclassified confidence scores as the number of weak learners increases. 
Other competitors are able to reduce the positive misclassfication confidence scores; however, their misclassfication confidence scores (are under the line) for the negative (majority) class are increasing e.g., CSB1, AdaC3. This highlights once more that the ability to adjust the weights during training is crucial to maintain good predictive performance across both classes. Note that for intermediate values of $T$, Figures are included in the Appendix as they depict this gradual behavior. 

An interesting observation is that the cost-sensitive methods become less confident in the confidence of the correctly classified instances (both classes) as the number of weak learners increases. As it seems, \textit{the more they learn}, their mistakes are reduced but they also \textit{become less confident} in their correct decisions.

\noindent\textbf{Decision Boundary Analysis:} Finally, we generate an imbalanced dataset similar to the toy dataset in Figure~\ref{fig:toy_example} of 40 instances (30 red class and 10 green class) with 2 features (for better visualization). We train each method on the same dataset and afterwards we show the decision boundaries which are learned from the training set. Since the dataset has only two features $x_1$ and $x_2$, we use a small number of weak learners ($T=5$). 
In Figure~\ref{fig:boundaries} we show the decision boundaries of all methods and how each method changes the weight distribution over the boosting rounds.

As we can see, AdaBoost gives more emphasis to the majority (red) class, while it tunes for overall classification accuracy. AdaCC1 and AdaCC2 on this particular dataset behave similarly by properly partitioning the space, giving emphasis to minority class without deteriorating the performance on majority class (2 blue misclassified points versus 4 red misclassified points). AdaMEC and AdaMEC-Cal. cannot find, through grid search, good misclassifications costs; therefore, their behavior is similar to AdaBoost (by considering the best $C_N = 1$ which makes them behave equal to AdaBoost). The misclassification cost selection of the competitors is based upon the performance of the final ensemble while our methods dynamically adapt their misclassification costs on each boosting round. CGAda, CSB1, CSB2 and AdaC2 partition the space to allow higher recall scores; however, they misclassify 12 red points. AdaCost, AdaC1 and AdaC3 perform even worse by misclassifying 19 red points. Interestingly, RareBoost partitions the space in a safe way e.g., it correctly classifies 5 blue points and the majority class.


%% file: conclusions.tex
In this work we present a novel strategy for cost-sensitive boosting that exploits the cumulative behavior of the model to dynamically balance the misclassification costs on each boosting round. 

Existing approaches require a user-defined fixed misclassification cost matrix as input. In most cases
this results in additional hyperparameters which need to be optimized jointly with the basic 
parameters, e.g., using grid search.
As grid-search does not ensure a good initialization it might hurt the model's overall predictive performance. Our methods' ability to produce
 consistent improvements in different measures, e.g., [0.3\%-28.56\%] for the AUC, [3.4\%-21.4\%] for the balanced accuracy, [4.8\%-45\%] for gmean and [7.4\%-85.5\%] for the recall indicate the general applicability of our method. The high recall scores demonstrate, that our method is especially helpful for domains in which low recall scores have a disastrous impact. 
Moreover, we have shown the superior performance of such cumulative models comparing to their non-cumulative counterparts, in terms of both predictive performance and model stability. Finally, our method comes with theoretical guarantees w.r.t. the training error and it reduces the optimization of hyper-parameters.

In the future, we will consider multi-class extensions of our method. Furthermore, we plan to investigate our method's application to the supervised online learning task. Our method's ability to dynamically adjust the misclassification costs, makes our method suitable for such a task in contrast to a recent online cost-sensitive boosting extension of AdaC2~\cite{wang2016}.  

%% file: appendix.tex
\begin{table}[b!]
\caption{Comparative Balanced Accuracy ranks across the entire set of methods and datasets (smaller values are better) for $T=25$. Best methods per dataset are in bold. Last row (winner) indicates on how many datasets a method is ranked first (best balanced accuracy score, higher values are better). Note that in some datasets, the methods have equal scores (tie); therefore, the ranks are in float format.}
\begin{adjustbox}{width=1\textwidth}
\begin{tabular}{lcccccccccccccc}
\hline
 & AdaBoost & AdaCC1 & AdaCC2 & AdaMEC & AdaMEC-Cal. & CGAda & CGAda-Cal. & AdaCost & CSB1 & CSB2 & AdaC1 & AdaC2 & AdaC3 & RareBoost \\ \hline
abalone & 14.0 & 3.0 & 5.0 & 2.0 & 9.0 & 8.0 & 10.5 & 4.0 & 10.5 & \textbf{1.0} & 6.0 & 7.0 & 12.0 & 13.0 \\
adult & 14.0 & \textbf{1.0} & 2.0 & 3.0 & 7.0 & 6.0 & 8.0 & 5.0 & 10.0 & 13.0 & 12.0 & 4.0 & 9.0 & 11.0 \\
bank & 14.0 & 2.0 & \textbf{1.0} & 8.0 & 3.0 & 5.0 & 4.0 & 9.0 & 11.0 & 6.0 & 12.0 & 10.0 & 7.0 & 13.0 \\
car eval. & 12.0 & \textbf{1.0} & 2.0 & 6.0 & 4.5 & 3.0 & 4.5 & 13.0 & 8.0 & 11.0 & 14.0 & 10.0 & 9.0 & 7.0 \\
coil 2000 & 14.0 & 5.0 & 2.0 & 7.0 & 10.0 & 9.0 & 6.0 & 11.0 & 12.0 & \textbf{1.0} & 4.0 & 8.0 & 3.0 & 13.0 \\
credit & 12.5 & 4.0 & 9.0 & 5.0 & \textbf{1.0} & 3.0 & 2.0 & 7.0 & 8.0 & 12.5 & 6.0 & 11.0 & 10.0 & 14.0 \\
eeg eye & 4.0 & 3.0 & 8.0 & 9.0 & 2.0 & 5.0 & 7.0 & 14.0 & 12.0 & 10.0 & 6.0 & 11.0 & 13.0 & \textbf{1.0} \\
electricity & 8.5 & \textbf{1.0} & 7.0 & 5.0 & 3.0 & 6.0 & 4.0 & 12.0 & 14.0 & 8.5 & 11.0 & 10.0 & 13.0 & 2.0 \\
isolet & 14.0 & 2.0 & \textbf{1.0} & 7.0 & 11.0 & 10.0 & 9.0 & 12.0 & 5.0 & 3.0 & 13.0 & 4.0 & 6.0 & 8.0 \\
letter img & 9.5 & \textbf{1.0} & 2.0 & 9.5 & 4.0 & 5.0 & 6.0 & 7.0 & 12.0 & 9.5 & 14.0 & 9.5 & 13.0 & 3.0 \\
mammography & 14.0 & 2.0 & \textbf{1.0} & 8.0 & 3.0 & 6.0 & 4.0 & 7.0 & 5.0 & 11.0 & 9.0 & 12.0 & 10.0 & 13.0 \\
musk2 & 13.0 & 2.0 & \textbf{1.0} & 6.0 & 8.0 & 5.0 & 7.0 & 9.0 & 4.0 & 13.0 & 11.0 & 13.0 & 10.0 & 3.0 \\
optical digits & 10.0 & 2.0 & \textbf{1.0} & 8.0 & 6.0 & 5.0 & 7.0 & 13.0 & 4.0 & 10.0 & 12.0 & 10.0 & 14.0 & 3.0 \\
ozone level & 14.0 & \textbf{1.0} & 4.0 & 7.0 & 9.0 & 11.0 & 10.0 & 6.0 & 8.0 & 2.0 & 12.0 & 3.0 & 5.0 & 13.0 \\
pen digits & 10.0 & 2.0 & \textbf{1.0} & 7.0 & 5.0 & 3.5 & 3.5 & 12.0 & 8.0 & 10.0 & 13.0 & 10.0 & 14.0 & 6.0 \\
phoneme & 13.0 & 7.0 & 6.0 & 4.0 & 3.0 & \textbf{1.0} & 2.0 & 11.0 & 8.0 & 12.0 & 10.0 & 5.0 & 14.0 & 9.0 \\
protein homo & 10.5 & \textbf{1.0} & 2.0 & 10.5 & 3.0 & 6.0 & 8.0 & 14.0 & 13.0 & 10.5 & 5.0 & 10.5 & 7.0 & 4.0 \\
satimage & 14.0 & \textbf{1.0} & 2.0 & 6.0 & 8.5 & 10.0 & 8.5 & 12.0 & 7.0 & 4.0 & 11.0 & 5.0 & 3.0 & 13.0 \\
scene & 14.0 & 3.0 & 4.0 & 7.0 & 11.0 & 8.0 & 10.0 & 5.0 & 9.0 & \textbf{1.0} & 6.0 & 2.0 & 12.0 & 13.0 \\
sick euthyroid & 10.0 & 2.0 & \textbf{1.0} & 6.0 & 7.5 & 4.0 & 7.5 & 13.0 & 5.0 & 10.0 & 3.0 & 10.0 & 12.0 & 14.0 \\
skin & 14.0 & 5.0 & \textbf{1.0} & 6.0 & 2.5 & 4.0 & 2.5 & 13.0 & 9.0 & 8.0 & 11.0 & 7.0 & 12.0 & 10.0 \\
spam & 10.5 & 4.0 & 2.0 & 10.5 & 5.5 & 3.0 & 5.5 & 14.0 & 13.0 & 10.5 & 7.0 & 10.5 & 8.0 & \textbf{1.0} \\
thyroid sick & 13.0 & \textbf{1.0} & 2.0 & 4.0 & 5.0 & 7.0 & 6.0 & 3.0 & 8.0 & 11.0 & 10.0 & 9.0 & 14.0 & 12.0 \\
us crime & 13.0 & \textbf{1.0} & 2.0 & 5.0 & 6.0 & 10.0 & 7.0 & 4.0 & 3.0 & 11.0 & 8.0 & 12.0 & 9.0 & 14.0 \\
webpage & 13.0 & \textbf{1.0} & 2.0 & 5.0 & 4.0 & 8.0 & 3.0 & 11.0 & 7.0 & 9.0 & 14.0 & 6.0 & 12.0 & 10.0 \\
wilt & 13.0 & 2.0 & \textbf{1.0} & 8.0 & 4.0 & 9.0 & 5.0 & 12.0 & 10.0 & 3.0 & 14.0 & 6.0 & 7.0 & 11.0 \\
wine quality & 14.0 & 2.0 & \textbf{1.0} & 4.0 & 6.0 & 9.0 & 7.0 & 10.0 & 8.0 & 3.0 & 11.0 & 5.0 & 12.0 & 13.0 \\ \hline
avg. & 12.20 & \textbf{2.30} & 2.70 & 6.43 & 5.61 & 6.28 & 6.09 & 9.74 & 8.57 & 7.94 & 9.81 & 8.17 & 10.00 & 9.15 \\ \hline
winner & 0 & \textbf{10} & \textbf{10} & 0 & 1 & 1 & 0 & 0 & 0 & 3 & 0 & 0 & 0 & 2 \\ \hline
\end{tabular}\end{adjustbox}
\end{table}

\begin{table}
\caption{Comparative Balanced Accuracy ranks across the entire set of methods and datasets (smaller values are better) for $T=50$. Best methods per dataset are in bold. Last row (winner) indicates on how many datasets a method is ranked first (best balanced accuracy score, higher values are better). Note that in some datasets, the methods have equal scores (tie); therefore, the ranks are in float format.}
\begin{adjustbox}{width=1\textwidth}
\begin{tabular}{lcccccccccccccc}
\hline
 & AdaBoost & AdaCC1 & AdaCC2 & AdaMEC & AdaMEC-Cal. & CGAda & CGAda-Cal. & AdaCost & CSB1 & CSB2 & AdaC1 & AdaC2 & AdaC3 & RareBoost \\ \hline
abalone & 14.0 & 2.0 & 3.0 & \textbf{1.0} & 8.0 & 7.0 & 9.0 & 4.0 & 10.0 & 12.0 & 5.0 & 6.0 & 11.0 & 13.0 \\
adult & 12.0 & 3.0 & \textbf{1.0} & 2.0 & 5.0 & 6.0 & 4.0 & 7.0 & 14.0 & 12.0 & 10.0 & 12.0 & 8.0 & 9.0 \\
bank & 13.5 & 2.0 & \textbf{1.0} & 6.0 & 5.0 & 3.5 & 3.5 & 8.0 & 9.0 & 13.5 & 11.0 & 10.0 & 7.0 & 12.0 \\
car eval. & 11.0 & 2.0 & \textbf{1.0} & 8.0 & 3.5 & 5.0 & 3.5 & 13.0 & 7.0 & 11.0 & 9.0 & 11.0 & 14.0 & 6.0 \\
coil 2000 & 14.0 & 3.0 & \textbf{1.0} & 9.0 & 8.0 & 6.0 & 4.0 & 7.0 & 10.0 & 13.0 & 2.0 & 5.0 & 11.0 & 12.0 \\
credit & 12.5 & 3.0 & 8.0 & 6.0 & \textbf{1.0} & 2.0 & 5.0 & 4.0 & 10.0 & 12.5 & 7.0 & 11.0 & 9.0 & 14.0 \\
eeg eye & 3.0 & 4.0 & 9.0 & 11.0 & 2.0 & 6.0 & 5.0 & 12.0 & 14.0 & 7.0 & 8.0 & 10.0 & 13.0 & \textbf{1.0} \\
electricity & 8.5 & 3.0 & 6.0 & 7.0 & 4.0 & 2.0 & 5.0 & 13.0 & 14.0 & 8.5 & 11.0 & 10.0 & 12.0 & \textbf{1.0} \\
isolet & 12.5 & \textbf{1.0} & 2.0 & 12.5 & 6.0 & 7.0 & 3.0 & 8.0 & 5.0 & 12.5 & 10.0 & 12.5 & 9.0 & 4.0 \\
letter img & 9.5 & \textbf{1.0} & 2.0 & 9.5 & 4.0 & 6.0 & 7.0 & 12.0 & 5.0 & 9.5 & 13.0 & 9.5 & 14.0 & 3.0 \\
mammography & 13.5 & 2.0 & \textbf{1.0} & 4.0 & 3.0 & 6.0 & 5.0 & 7.0 & 9.0 & 13.5 & 8.0 & 12.0 & 10.0 & 11.0 \\
musk2 & 11.0 & 3.0 & 2.0 & 9.0 & 6.0 & 5.0 & 7.0 & 8.0 & 4.0 & 11.0 & 13.0 & 11.0 & 14.0 & \textbf{1.0} \\
optical digits & 9.5 & 2.0 & \textbf{1.0} & 9.5 & 5.5 & 5.5 & 4.0 & 13.0 & 7.0 & 9.5 & 12.0 & 9.5 & 14.0 & 3.0 \\
ozone level & 13.0 & \textbf{1.0} & 4.0 & 2.0 & 5.0 & 7.0 & 6.0 & 8.0 & 9.0 & 13.0 & 11.0 & 13.0 & 3.0 & 10.0 \\
pen digits & 9.5 & 2.0 & \textbf{1.0} & 9.5 & 6.0 & 4.0 & 5.0 & 12.0 & 7.0 & 9.5 & 13.0 & 9.5 & 14.0 & 3.0 \\
phoneme & 9.5 & 6.0 & \textbf{1.0} & 4.0 & 3.0 & 2.0 & 5.0 & 13.0 & 12.0 & 9.5 & 11.0 & 7.0 & 14.0 & 8.0 \\
protein homo & 9.5 & 2.0 & \textbf{1.0} & 9.5 & 3.0 & 5.0 & 6.0 & 14.0 & 13.0 & 9.5 & 7.0 & 9.5 & 12.0 & 4.0 \\
satimage & 14.0 & 2.0 & \textbf{1.0} & 10.0 & 7.0 & 5.5 & 5.5 & 11.0 & 4.0 & 13.0 & 8.0 & 9.0 & 3.0 & 12.0 \\
scene & 14.0 & \textbf{1.0} & 3.0 & 9.0 & 7.0 & 8.0 & 6.0 & 4.0 & 10.0 & 11.0 & 5.0 & 2.0 & 12.0 & 13.0 \\
sick euthyroid & 9.5 & 2.0 & \textbf{1.0} & 9.5 & 7.0 & 4.0 & 6.0 & 13.0 & 12.0 & 9.5 & 3.0 & 9.5 & 5.0 & 14.0 \\
skin & 12.5 & \textbf{1.0} & 6.0 & 2.0 & 4.5 & 3.0 & 4.5 & 14.0 & 9.0 & 12.5 & 10.0 & 7.0 & 11.0 & 8.0 \\
spam & 8.5 & 6.0 & 2.0 & 8.5 & 4.5 & 3.0 & 4.5 & 14.0 & 13.0 & 8.5 & 11.5 & 8.5 & 11.5 & \textbf{1.0} \\
thyroid sick & 12.5 & 2.0 & \textbf{1.0} & 6.0 & 4.0 & 7.0 & 5.0 & 3.0 & 8.0 & 12.5 & 9.0 & 11.0 & 14.0 & 10.0 \\
us crime & 12.5 & \textbf{1.0} & 2.0 & 6.0 & 4.0 & 9.0 & 7.0 & 3.0 & 5.0 & 12.5 & 8.0 & 11.0 & 10.0 & 14.0 \\
webpage & 12.5 & 3.0 & \textbf{1.0} & 2.0 & 6.0 & 9.0 & 7.0 & 8.0 & 5.0 & 12.5 & 14.0 & 10.0 & 11.0 & 4.0 \\
wilt & 10.0 & 2.0 & \textbf{1.0} & 6.0 & 3.0 & 5.0 & 4.0 & 12.0 & 7.0 & 10.0 & 13.0 & 10.0 & 14.0 & 8.0 \\
wine quality & 14.0 & \textbf{1.0} & 2.0 & 8.0 & 3.0 & 5.0 & 4.0 & 7.0 & 6.0 & 13.0 & 10.0 & 9.0 & 11.0 & 12.0 \\ \hline
avg. & 11.33 & \textbf{2.33} & 2.41 & 6.91 & 4.74 & 5.31 & 5.20 & 9.33 & 8.81 & 11.19 & 9.35 & 9.46 & 10.80 & 7.81 \\ \hline
winner & 0 & 7 & \textbf{14} & 1 & 1 & 0 & 0 & 0 & 0 & 0 & 0 & 0 & 0 & 4 \\ \hline
\end{tabular}\end{adjustbox}
\end{table}

\begin{table}
\caption{Comparative Balanced Accuracy ranks across the entire set of methods and datasets (smaller values are better) for $T=100$. Best methods per dataset are in bold. Last row (winner) indicates on how many datasets a method is ranked first (best balanced accuracy score, higher values are better). Note that in some datasets, the methods have equal scores (tie); therefore, the ranks are in float format.}
\begin{adjustbox}{width=1\textwidth}
\begin{tabular}{lcccccccccccccc}
\hline
 & AdaBoost & AdaCC1 & AdaCC2 & AdaMEC & AdaMEC-Cal. & CGAda & CGAda-Cal. & AdaCost & CSB1 & CSB2 & AdaC1 & AdaC2 & AdaC3 & RareBoost \\ \hline
abalone & 14.0 & 2.0 & 3.0 & 9.0 & 6.0 & 7.0 & 8.0 & \textbf{1.0} & 11.0 & 13.0 & 4.0 & 5.0 & 10.0 & 12.0 \\
adult & 12.0 & \textbf{1.0} & 2.0 & 3.0 & 4.0 & 6.0 & 5.0 & 9.0 & 14.0 & 12.0 & 10.0 & 12.0 & 8.0 & 7.0 \\
bank & 12.5 & \textbf{1.0} & 2.0 & 3.0 & 6.0 & 4.0 & 5.0 & 8.0 & 14.0 & 12.5 & 10.0 & 9.0 & 7.0 & 11.0 \\
car eval. & 9.5 & 2.0 & \textbf{1.0} & 9.5 & 3.0 & 5.0 & 4.0 & 13.0 & 7.0 & 9.5 & 12.0 & 9.5 & 14.0 & 6.0 \\
coil 2000 & 13.0 & \textbf{1.0} & 2.0 & 8.0 & 7.0 & 4.0 & 6.0 & 9.0 & 10.0 & 14.0 & 3.0 & 5.0 & 12.0 & 11.0 \\
credit & 12.5 & 3.0 & 8.0 & \textbf{1.0} & 2.0 & 4.0 & 5.0 & 6.0 & 11.0 & 12.5 & 7.0 & 10.0 & 9.0 & 14.0 \\
eeg eye & 4.5 & 3.0 & 10.0 & 11.0 & 2.0 & 8.0 & 6.0 & 14.0 & 13.0 & 4.5 & 9.0 & 7.0 & 12.0 & \textbf{1.0} \\
electricity & 8.0 & 3.0 & 6.0 & 10.0 & 4.0 & 2.0 & 5.0 & 13.0 & 14.0 & 8.0 & 11.0 & 8.0 & 12.0 & \textbf{1.0} \\
isolet & 10.5 & \textbf{1.0} & 2.0 & 10.5 & 6.0 & 4.0 & 5.0 & 13.0 & 8.0 & 10.5 & 7.0 & 10.5 & 14.0 & 3.0 \\
letter img & 8.5 & 2.0 & \textbf{1.0} & 8.5 & 4.0 & 5.0 & 6.0 & 12.0 & 11.0 & 8.5 & 13.5 & 8.5 & 13.5 & 3.0 \\
mammography & 12.0 & 2.0 & \textbf{1.0} & 4.0 & 3.0 & 5.0 & 6.0 & 7.0 & 14.0 & 12.0 & 8.0 & 12.0 & 10.0 & 9.0 \\
musk2 & 9.5 & 3.0 & 2.0 & 9.5 & 5.5 & 7.0 & 5.5 & 12.0 & 4.0 & 9.5 & 13.0 & 9.5 & 14.0 & \textbf{1.0} \\
optical digits & 9.5 & 2.0 & \textbf{1.0} & 9.5 & 4.0 & 6.0 & 5.0 & 14.0 & 7.0 & 9.5 & 12.0 & 9.5 & 13.0 & 3.0 \\
ozone level & 13.0 & \textbf{1.0} & 2.0 & 9.0 & 4.0 & 7.0 & 6.0 & 5.0 & 8.0 & 13.0 & 11.0 & 13.0 & 3.0 & 10.0 \\
pen digits & 8.5 & 2.0 & \textbf{1.0} & 8.5 & 6.0 & 4.0 & 5.0 & 14.0 & 11.0 & 8.5 & 12.0 & 8.5 & 13.0 & 3.0 \\
phoneme & 9.5 & 6.0 & \textbf{1.0} & 5.0 & 2.0 & 4.0 & 3.0 & 12.0 & 14.0 & 9.5 & 11.0 & 7.0 & 13.0 & 8.0 \\
protein homo & 8.5 & 2.0 & \textbf{1.0} & 8.5 & 3.0 & 5.0 & 6.0 & 14.0 & 12.0 & 8.5 & 11.0 & 8.5 & 13.0 & 4.0 \\
satimage & 13.0 & 2.0 & \textbf{1.0} & 11.0 & 6.0 & 4.0 & 5.0 & 8.0 & 10.0 & 13.0 & 7.0 & 13.0 & 3.0 & 9.0 \\
scene & 13.5 & \textbf{1.0} & 6.0 & 3.0 & 7.0 & 9.0 & 8.0 & 2.0 & 10.0 & 13.5 & 4.0 & 5.0 & 12.0 & 11.0 \\
sick euthyroid & 9.5 & 2.0 & \textbf{1.0} & 9.5 & 7.0 & 5.0 & 6.0 & 12.0 & 14.0 & 9.5 & 3.0 & 9.5 & 4.0 & 13.0 \\
skin & 8.5 & 3.0 & 4.0 & 8.5 & \textbf{1.5} & 5.0 & \textbf{1.5} & 13.0 & 14.0 & 8.5 & 11.0 & 8.5 & 12.0 & 6.0 \\
spam & 8.5 & 4.0 & 2.0 & 8.5 & 5.5 & 3.0 & 5.5 & 14.0 & 13.0 & 8.5 & 11.5 & 8.5 & 11.5 & \textbf{1.0} \\
thyroid sick & 12.5 & 2.0 & \textbf{1.0} & 7.0 & 4.0 & 6.0 & 5.0 & 3.0 & 9.5 & 12.5 & 9.5 & 11.0 & 14.0 & 8.0 \\
us crime & 12.0 & \textbf{1.0} & 3.0 & 10.0 & 5.0 & 7.0 & 6.0 & 2.0 & 4.0 & 12.0 & 8.0 & 12.0 & 14.0 & 9.0 \\
webpage & 12.5 & 2.0 & \textbf{1.0} & 7.0 & 4.0 & 6.0 & 5.0 & 8.0 & 9.0 & 12.5 & 14.0 & 10.0 & 11.0 & 3.0 \\
wilt & 9.5 & 2.0 & \textbf{1.0} & 9.5 & 3.0 & 5.0 & 4.0 & 12.0 & 7.0 & 9.5 & 13.0 & 9.5 & 14.0 & 6.0 \\
wine quality & 13.5 & \textbf{1.0} & 2.0 & 4.0 & 3.0 & 7.0 & 6.0 & 5.0 & 10.0 & 13.5 & 9.0 & 8.0 & 11.0 & 12.0 \\ \hline
avg. & 10.69 & \textbf{2.11} & 2.52 & 7.61 & 4.35 & 5.33 & 5.31 & 9.44 & 10.50 & 10.69 & 9.43 & 9.17 & 11.00 & 6.85 \\ \hline
winner & 0 & 8 & \textbf{12} & 1 & 0.5 & 0 & 0.5 & 1 & 0 & 0 & 0 & 0 & 0 & 4 \\ \hline
\end{tabular}\end{adjustbox}
\end{table}

\begin{figure*}[htp!]
 \centering
 \centering
 \begin{subfigure}[t]{1\textwidth}
 \centering
 \includegraphics[width=1\textwidth]{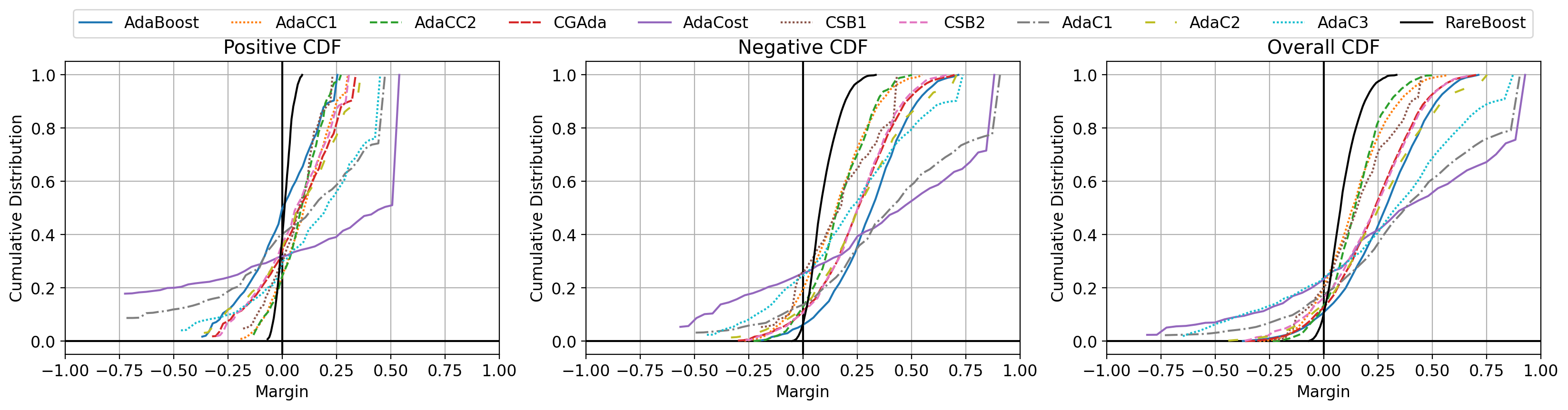}
 \caption{T = 50}
 \label{fig:cdf_t_50_appendix}
 \end{subfigure}
 \begin{subfigure}[t]{1\textwidth}
 \centering
 \includegraphics[width=1\textwidth]{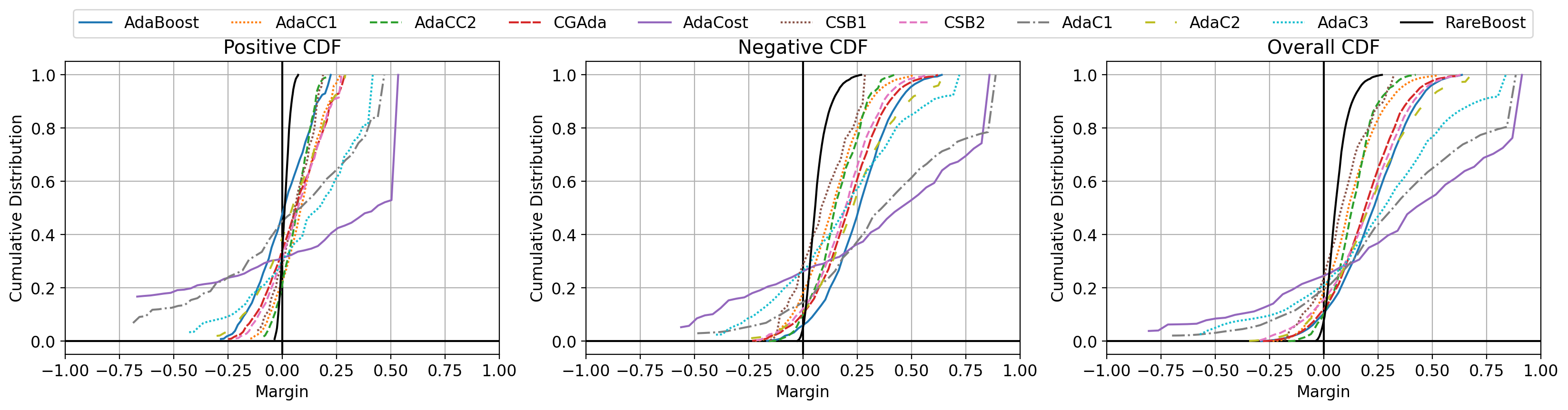}
 \caption{T = 100}
 \label{fig:cdf_t_100_appendix}
 \end{subfigure}
 \caption{Effect of boosting rounds on the confidence scores (left:positive class, middle:negative class, left:overall).}
 \label{fig:cdf_plots_appendix}
\end{figure*}